\documentclass[11pt]{article}

\usepackage[pdftex]{graphicx}
\usepackage{hyperref}
\usepackage{amssymb}
\usepackage{amsmath}
\usepackage{bm}
\usepackage{mathtools}
\usepackage{booktabs}
\usepackage{algorithm}
\usepackage[noend]{algorithmic}

\title{\vspace*{-0.75in}\bf From Prediction to Prescription: Evolutionary Optimization of Non-Pharmaceutical Interventions in the COVID-19 Pandemic}
\author{Risto Miikkulainen$^{1,2}$, Olivier Francon$^1$, Elliot Meyerson$^1$,\\
Xin Qiu$^1$, Elisa Canzani$^1$, and Babak Hodjat$^1$\\[1ex]
$^1$Cognizant Technology Solutions and $^2$The University of Texas at Austin
}
\small\normalsize
\date{}
\begin{document}

\maketitle

\begin{abstract}
Several	models have been developed to predict how the COVID-19
pandemic spreads, and how it could be contained with
non-pharmaceutical interventions (NPIs) such as social distancing
restrictions and school and business closures. This paper demonstrates
how evolutionary AI could be used to facilitate the next step, i.e.\
determining most effective intervention strategies automatically.
Through evolutionary surrogate-assisted prescription (ESP), it is
possible to generate a large number of candidate strategies and
evaluate them with predictive models.  In principle, strategies can be
customized for different countries and locales, and balance the need
to contain the pandemic and the need to minimize their economic
impact. While still limited by available data, early experiments
suggest that workplace and school restrictions are the most important
and need to be designed carefully. It also demonstrates
that results of lifting restrictions can be unreliable, and suggests
creative ways in which restrictions can be implemented softly, e.g.\
by alternating them over time. As more data becomes available, the
approach can be increasingly useful in dealing with COVID-19 as well
as possible future pandemics.
\end{abstract}

\section{Introduction}

The COVID-19 crisis is unprecedented in modern times, and caught the
world largely unprepared. Since there is little experience and
guidance, authorities have been responding in a variety of ways.
Many different non-pharmaceutical interventions (NPIs) have been
implemented at different stages of the pandemic and in different
contexts. On the other hand, compared to past pandemics, for the first
time almost real-time data is collected about these interventions,
their economic impact, and the spread of the disease. These two
factors create an excellent opportunity for computational modeling and
machine learning.

Most of the modeling efforts so far have been based on traditional
epidemiological methods, such as compartmental models \cite{cdc2020}.
Such models can be used to predict the spread of the disease, assuming
a few parameters such as the basic reproduction number $R_0$ can be estimated
accurately. New ideas have also emerged, including using cell-phone
data to measure social distancing \cite{Woody:UT2020}.  These models have
been extended with NPIs by modifying the transmission rates: each NPI
is assumed to reduce the transmission rate by a certain amount
\cite{stanford2020,Murray:IHME2020,flaxman:imperial2020}. Such models
have received a lot of attention: in
this unprecedented situation, they are our only source of support for
making informed decisions on how to reduce and contain the spread of the disease.

However, epidemiological models are far from perfect. Much about how
the disease is transmitted, how prevalent it is in the population, how
many people are immune and the strength of the immunity, is unknown, and
it is difficult to parameterize the models accurately. Similarly, the
effects of NPIs are unpredictable in that their effect varies based on
the cultural and economic environment, the stage of the pandemic, and
above all, they interact in nonlinear ways. To overcome the
uncertainty, data is crucial. Instead of estimating parameters, it is
possible to fit the models to the data so that they predict already
existing data more accurately. In the extreme, with enough data, it is
possible to use machine learning simply to model the data. All the
unknown epidemiological, cultural, and economic parameters and
interactions are expressed in the time series of infections and
NPIs. The fact that the epidemic spreads to different regions at a lag, and different geographies experience different stages of the spread at different times, and often react to it differently, gives us the opportunity to 'front-run' the epidemic with data-driven models. Machine learning can then be used to construct a model, such as
a recurrent neural network, that accurately predicts the outcomes
without having to understand precisely how they emerge.

The data-driven modeling approach is implemented and evaluated in this
paper. An LSTM neural network model \cite{hochreiter:nc97,greff:ieeetnn17}
is trained with
publicly available data on infections and NPIs \cite{hale:data20} in a
number of countries and applied to predicting how the pandemic will unfold in
them in the future. The predictions are cascaded one day at a time and
constrained to a meaningful range. Even with current limited data, the
predictions are surprisingly accurate and well-behaved.  This result
suggests that the data-driven machine learning approach is potentially
a powerful new tool for epidemiological modeling. This is the first
main contribution of the paper.

A more significant contribution, however, is to demonstrate that machine
learning can also be used to take the next step, i.e.\ to extend the
models from prediction to prescription. That is, given that we can
predict how the NPIs affect the pandemic, we can also automatically
discover effective NPI strategies. The technology required for this
step is different from standard machine learning. The goal is not to
model and predict processes for which data already exists, but to create new
solutions that may have never existed before.  In other words, it
requires extending AI from imitation to creativity.

\begin{figure}
    \centering
    \includegraphics[width=0.3\textwidth]{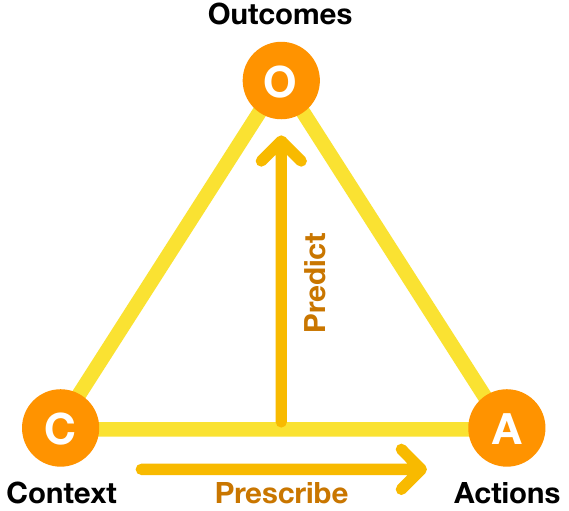}
    \caption{Elements of the ESP Decision Optimization Method.
      A Predictor is trained with historical data on how given actions          
      in given contexts led to specific outcomes. For instance in the           
      NPI optimization problem, given the state of pandemic in a                
      particular country and the NPIs in effect, it predicts the                
      future number of cases and deaths and the cost of the NPIs.  The          
      Predictor can be any machine learning model trained with                  
      supervised methods, such as a random forest or a neural network,          
      or even a simulation such as an epidemiological model. The                
      Predictor is then used as a surrogate in order to evolve a                
      Prescriptors, i.e.\ neural networks that implement decision               
      policies (i.e.\ NPIs) resulting in best possible outcomes.  With           
      multiple conflicting objectives (such as cases/deaths and cost),          
      evolution results in multiple Prescriptors, each representing a           
      different tradeoff, from which a human decision maker can choose          
      the ones that best matches their preferences. }
    \label{fg:esptriangle}
\end{figure}

This extension is indeed underway in AI through several approaches
such as reinforcement learning, Bayesian parameter optimization,
gradient-based approaches, and evolutionary computation
\cite{miikkulainen:action18,elsken:jmlr19,lehman:alife20}. 
The approach taken in this
paper is based on evolutionary surrogate-assisted prescription
\cite{francon:gecco20}, a technique that combines evolutionary search with
surrogate modeling (Figure~\ref{fg:esptriangle}).

In ESP, a predictive model is first formed through standard machine
learning techniques, such as neural networks. Given actions taken in a
given context (such as NPIs at a given stage of the pandemic), it
predicts what the outcomes would be (such as infections, deaths, and
economic cost).  A prescriptive model, another neural network, is then
formed to implement an optimal decision strategy, i.e.\ what actions
should be taken in each context. Since optimal actions are not known,
the Prescriptor cannot be trained with standard supervised
learning. However, it can be evolved, i.e.\ discovered through
population-based search. Because it is often impossible or
prohibitively costly to evaluate each candidate strategy in the real
world, the Predictor model is used as a surrogate. In this manner,
millions of candidate strategies can be generated and tested in order
to find ones that optimize the desired outcomes.

ESP has been used in several real-world design and decision
optimization problems, including discovering growth recipes for
agriculture \cite{johnson:plos19}, and designs for e-commerce websites
\cite{miikkulainen:aimag20}. It often discovers effective solutions that are
overlooked by human designers. Recently it has also been applied to
sequential decision-making tasks, and found to be more
sample-efficient, reliable, and safe than standard reinforcement
learning techniques \cite{francon:gecco20}. Much of this performance is due to a
surprising regularization effect that incompletely trained Predictors
and Prescriptors bring about.

The ESP approach is applied in this paper to the problem of
determining optimal NPIs for the COVID-19 pandemic. Using the
data-driven LSTM model as the Predictor, a Prescriptor is evolved in a
multi-objective setting to minimize the number of COVID-19 cases, as
well as the number and stringency of NPIs (representing economic
impact). In this process, evolution discovers a Pareto front of
Prescriptors that represent different tradeoffs between these two
objectives: Some utilize many NPIs to bring down the number of cases,
and others minimize the number of NPIs with a cost of more
cases. Therefore, the AI system is not designed to replace human
decision makers, but instead to empower them: Humans choose which
tradeoffs are the best, and the AI makes suggestions on how they can
be achieved. It therefore constitutes a step towards using AI not just
to model the pandemic, but to help contain it---which has so far been
missing from the literature \cite{darby:wp20}.

The current implementation should be taken as a proof of concept i.e.\
a demonstration of the potential power of the approach. The currently
available data is still limited in quantity, accuracy, and detail. It
is not yet possible to draw specific prescriptions reliably, such that
e.g.\ in a particular country, a particular NPI can be safely lifted
on a particular date.  The results so far suggest that such
prescriptions may become possible in the next few months, as the
quality and quantity of the data improves. The experiments already
point to two general conclusions. First, school and workplace closings
turn out to be the two most important NPIs in the simulations: they
have the largest and most reliable effects on the number of cases
compared to e.g.\ restrictions on gatherings and travel.  Second,
partial or alternating NPIs may be effective. Prescriptors repeatedly turn certain NPIs on and
off over time, for example, schools opening and closing on a weekly basis seems to imply the need for restricting schools to be opened fewer days per week. This is a creative and surprising solution, given the
limited variability of NPIs that is available to the Prescriptors.
Together these conclusions already suggest a possible focus for
efforts on establishing and lifting NPIs, in order to achieve maximum
effect and minimal cost.

The paper begins with a background of epidemiological modeling and the
general ESP approach. The datasets used, the data-driven predictive
model developed, and the evolutionary optimization of the Prescriptor
will then be described. Illustrative results will be reviewed in
several cases, drawing general conclusions about the potential power
of the approach. Future work on utilizing epidemiological models,
supporting interactive exploration, modeling uncertainty, and
generating explainable prescriptions will be discussed.

An interactive demo, allowing users to explore the Pareto front of
Prescriptors on multiple countries, is available at
\url{https://evolution.ml/esp/npi}.

\section{Background: Modeling Epidemics}
\label{sc:background}

Different types of epidemiological models are briefly characterized,
followed by a review of existing COVID-19 modeling efforts, and the
emerging opportunity for machine learning models.

\subsection{Types of Epidemiological Models}

Modern epidemic modeling started with the compartmental SIR model
developed by McKendrick and Kermack at the beginning of the 20th
century \cite{kermack1927contribution}. The SIR model assumes
susceptible individuals (S) can get infected (I) and, after a certain
infectious period, die or recover (R), becoming immune afterwards.
The model then describes global transmission dynamics at a population
scale as a system of differential equations in continuous
time. Depending on disease characteristics, these compartments and the
flow patterns between them can be further refined: For instance in the
case of HIV, mixing and contact depends on age groups
\cite{may1987transmission}. The main limitation of such metapopulation
models is that random mixing is limited between individuals within
population subgroups, i.e.\ compartments.

Newman \cite{newman2002spread} showed that contact patterns can be
represented more accurately through a network topology, taking into
account geography, demographics, and social factors, and thus
overcoming limitations of compartmental models. Keeling and Eames
\cite{keeling2005networks} further highlighted the stochastic nature
of transmissions by demonstrating SIR epidemics on five types of social
and spatial networks for a population of 10,000 individuals.  Later
studies focused on evolutionary and adaptive networks
\cite{gross2006epidemic}, aiming to model the dynamics of social
links, such as frequency, intensity, locality, and duration of
contacts, which strongly influence the long term impacts of epidemics.

Indeed, it is now widely recognized that multiple factors at different
levels influence how epidemics spread \cite{swarup2014computational}.
Models have become more detailed and sophisticated, relying on
extensive computational power now available to simulate them.  In
addition to compartmental and network models, agent-based simulations
have emerged as a third simulation paradigm. Agent-based approaches
describe the overall dynamics of infection as result of events and
activities involving single individuals \cite{venkatramanan2018using},
resulting in potentially detailed but computationally demanding
processes.  For a more detailed review of the literature on
epidemiological models and their mathematical insights see
\cite{canzani2015insights,luke2012systems}.

\subsection{Modeling the COVID-19 Pandemic}

A variety of epidemiological simulations are currently used to model
the COVID-19 pandemic.  The focus is on simulating effects of NPIs
in order to support decision making about response policies.

For instance, Stanford University \cite{stanford2020} extended the SIR
model up to nine compartments including susceptible, exposed,
asymptomatic, presymptomatic, mild and severe symptomatic,
hospitalized, deceased, and recovered populations, as well as a
stochastic simulator for transitions between compartments, calibrated
with MIDAS data \cite{midas2020}. Currently, the simulator supports up
to three interventions at different times. In contrast, NPIs are
implemented in a more granular fashion in the Bayesian inference model
of Imperial College London \cite{flaxman:imperial2020}.  Their model
parameters are estimated empirically from ECDC data \cite{ecdc2020}
for 11 countries.  Deaths are then predicted as a function of
infections based on the distribution of average time between infections
and time-varying reproduction number $R_t$.

The Institute for Health Metrics and Evaluation (IHME) of Washington
University \cite{Murray:IHME2020} combined compartmental models and
mixed-effects nonlinear regression to predict cumulative and daily
death rate as a function of NPIs.  They also forecast health service
needs using a micro-simulation model. The IHME dashboard
\cite{ihme2020dash} compares estimated and confirmed daily infections
upon testing rate for both U.S. and European countries up to state-
and regional levels. Similarly, the University of Texas
\cite{Woody:UT2020} developed a statistical model based on nonlinear
Bayesian hierarchical regression with a negative-binomial model for
daily variation in death rates. The novelty is to estimate social
distancing using geolocation data from mobile phones, improving
accuracy over the IHME model. So far, the forecasts rely on U.S. data
and are not reliable beyond about weeks.

More broadly, the Centers for Disease Control and Prevention (CDC)
\cite{cdc2020} provides details about the different COVID-19
prediction models and their specific NPI assumptions. In fact, CDC
works with partners to bring together forecasts for cumulative deaths
over the following four weeks. Forecasting teams predict numbers of
deaths using different types of data (e.g.\ WHO, COVID-19,
demographic, mobility), methods, and estimates of the impacts of NPIs
(e.g.\ social distancing and use of face coverings). In general, most
of COVID-19 forecast approaches use curve-fitting and ensembles of
mechanistic SIR models with age structures and different parameter
assumptions. Social distancing and NPIs are usually not represented
directly, but approximated as changes in transmission rates. The main
advantage is that running simulations require a few input parameters
based on average data at population scale.

In contrast, since contact dynamics in agent-based and network
approaches results from events and activities of single individuals
and their locations, they can be more accurate in modeling social
distancing and NPIs. However, their parameters need to be calibrated
appropriately, which is difficult to do with available data.
Mechanistic transmission models can help overcome data collection
challenges when tracing a real network of individuals. Different
tracing techniques, such as comprehensive diary-based studies (e.g.\
POLYMOD \cite{mossong2008social}), and recorded movements of
individuals (e.g.\ transportation networks \cite{hufnagel2004forecast}
and dollar bill tacking \cite{brockmann2006scaling}) have been
investigated in the past to sample real networks with limited
resources and data availability. In the context of COVID-19, as
mentioned above, a new opportunity is to use mobile phone data to
support implementation of social distancing measures
(e.g.\ \cite{Woody:UT2020}). Debates on value and ethics of tracking
people movements to monitor the COVID-19 are still ongoing
\cite{dubov2020value}. Another challenge for current epidemiological
modeling methods is that they require significant computing resources
and sophisticated parallelization algorithms.  As a step to making
them feasible, EpiFast \cite{bisset2009epifast} reduces the SIR
epidemics simulation problem to a sequence of graph operations on
distributed memory systems, and thereby significantly decreases the
simulation cost. The COVID-19 pandemic has accelerated efforts to
develop solutions to these challenges, and is likely to result in
improved models in the future.

\subsection{Opportunity for Machine Learning Models}

Any of the above models that include NPI effects and generate
long-term predictions could be used as the Predictor with ESP, even
several of them together as an ensemble. Taking advantage of them is
indeed a compelling direction of future work
(Section~\ref{sc:future}).  However, this paper focuses on evaluating
a new opportunity: Building the model solely based on past data using
machine learning. Given that data about the COVID-19 pandemic is
generated, recorded, and made available more than any epidemic before, such a
new approach may be feasible for the first time.

There is a lot of promise in this data-driven approach. The
epidemiological models require several assumptions about the
population, culture, and environment, depend on several parameters
that are difficult to set accurately, and cannot take into account
many possible nonlinear and dynamic interactions between the NPIs, and
in the population. In contrast, all such complexities are implicitly
included in the data. The data-driven models are phenomenological,
i.e.\ they do not explain how the given outcomes are produced, but
given enough data, they can be accurate in predicting them. This is
the first hypotheses tested in this paper; as shown in
Section~\ref{sc:predictive}, it turns out that even with the limited
data available at this point, data-driven models can be useful.

All the models discussed so far are predictive: Based on knowledge of
the populations and the epidemic, and the data so far about its
progress in different populations and efforts to contain it, they
estimate how the disease will progress in the future. By themselves,
these models do not make recommendations, or prescriptions, of what
NPIs would be most effective. It is possible to manually set up hypothetical
future NPI strategies and use the models to evaluate how well they would
work. However, only a few strategies can be tested in this manner, and
the process is limited by the ability of human experts to think of
promising strategies. Given past experience with surrogate modeling
and population-based search, automated methods may be more effective
in this process. A method for doing so, ESP, will be described next.

\section{Method: Evolutionary Surrogate-assisted Prescription (ESP)}

ESP is a continuous black-box optimization process for adaptive
decision-making \cite{francon:gecco20}.  In ESP, a model of the
problem domain is used as a surrogate for the problem itself. This
model, called the Predictor ($P_d$), takes a decision as its input,
and predicts the outcomes of that decision.  A decision consists of a
context (i.e.\ a problem) and actions to be taken in that context.

A second model, called the Prescriptor ($P_s$), is then created. It
takes a context as its input, and outputs actions that would optimize
outcomes in that context. In order to develop the Prescriptor, the
Predictor is used as the surrogate, i.e.\ a less costly and risky
alternative to the real world.

More formally, given a set of possible contexts $\mathbb{C}$ and
possible actions $\mathbb{A}$, a decision policy $D$ returns a set of
actions $A$ to be performed in each context $C$:
\begin{equation}                                                                
  D(C) = A\;,                                                                   
\end{equation}
where $C \in \mathbb{C}$ and $A \in \mathbb{A}$. For each such $(C,A)$
pair there is a set of outcomes $O(C,A)$, and the Predictor $P_d$ is
defined as
\begin{equation}                                                                
P_d (C, A) = O,                                                                 
\end{equation}
and the Prescriptor $P_s$ implements the decision policy as
\begin{equation}                                                                
P_s (C) = A\;,                                                                  
\end{equation}
such that $\sum_{i,j} O_j(C_i,A_i)$ over all possible contexts $i$ 
and outcome dimensions $j$ is maximized (assuming they improve with
increase). It thus approximates the optimal decision policy for the
problem. Note that the optimal actions $A$ are not known, and must
therefore be found through search.

In the case of the NPI optimization problem, context $C$ consists of
information regarding a region. This might include data on the number
of available ICU beds, population distribution, time since the first
case of the disease, current COVID-19 cases, and fatality
rate. Actions $A$ in this case specify whether or not the different
possible NPIs are implemented within that region. The outcomes $O$ for
each decision measure the number of cases and fatalities within two
weeks of the decision, and the cost of each NPI.

\begin{figure}
    \centering
    \includegraphics[width=0.5\textwidth]{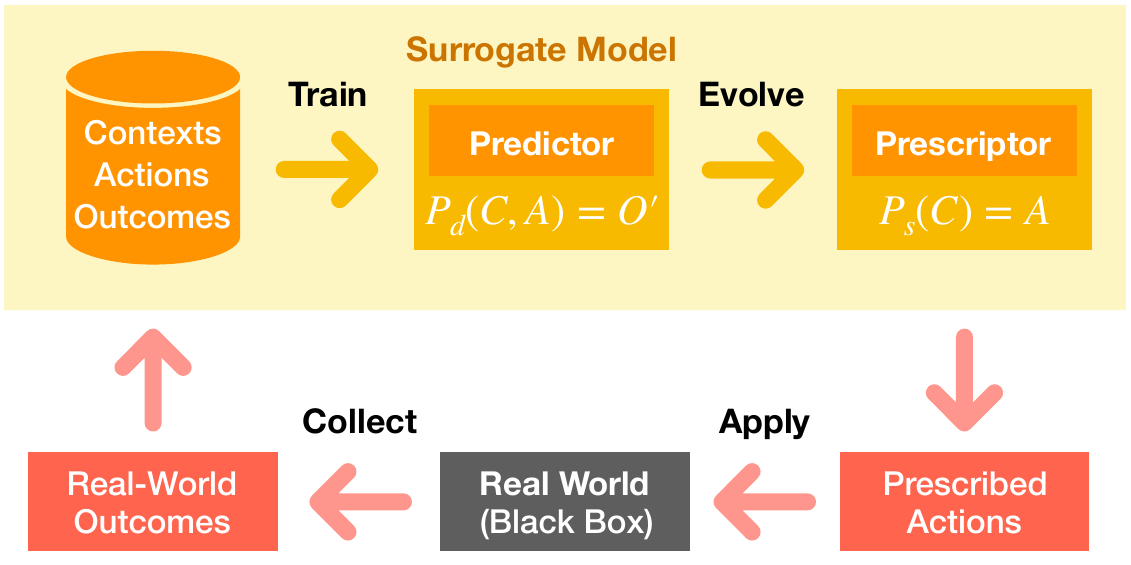}
    \caption{The ESP Outer Loop. The Predictor can be trained
      gradually at the same time as the Prescriptor is evolved, using           
      the Prescriptor to drive exploration. That is, the user can               
      decide to apply the Prescriptor's outputs to the real world,              
      observe the outcomes, and aggregate them into the Predictor's             
      training set. However, any new prescriptions implemented in the           
      real world, whether similar to the Prescriptors or not, can be            
      used to augment the training dataset. }
    \label{fg:outerloop}
\end{figure}

The ESP algorithm then operates as an outer loop that constructs the Predictor
and Prescriptor models (Figure~\ref{fg:outerloop}):
\begin{enumerate}
\setlength{\itemsep}{0ex}
\item\label{st:predictor} Train a Predictor based on historical training data;
\item Evolve Prescriptors with the Predictor as the surrogate;
\item Apply the best Prescriptor in the real world;
\item\label{st:newdata} Collect the new data and add to the training set;
\item Repeat.
\end{enumerate}
In the case of NPI optimization, there is currently no Step~3 since
the system is not yet incorporated into decision making. However, any
NPIs implemented in the real world, whether similar or dissimilar to
ESP's prescriptions, will similarly result in new training data. As
usual in evolutionary search, the process terminates when a
satisfactory level of outcomes has been reached, or no more progress
can be made, or the system iterates indefinitely, continuously adapting to changes in the real world (e.g., adapting to the advent of vaccines or antiviral drugs).

The Predictor model is built by modeling a $(C,A,O)$ dataset. The
choice of algorithm depends on the domain, i.e.\ how much data there
is, whether it is continuous or discrete, structured or unstructured.
Random forests, symbolic regression, and neural networks have been
used successfully in this role in the
past\cite{francon:gecco20,johnson:plos19}. In some cases, such as
NPI optimization, an ensemble of data-driven and simulation models
may be useful, in order to capture expected or fine-grained behavior
that might not yet have been reflected in the data
(Section~\ref{sc:future}).

The Prescriptor model is built using neuroevolution: Neural networks
because they can express complex nonlinear mappings naturally, and
evolution because it is an efficient way to discover such mappings
\cite{stanley:naturemi19}, and naturally suited to optimize multiple
objectives \cite{coello:ijkais99,emmerich:natcomp18}. Because it is
evolved with the Predictor, the Prescriptor is not restricted by a
finite training dataset, or limited opportunities to evaluate in the
real world. Instead, the Predictor serves as a fitness function, and
it can be queried frequently and efficiently.  In a multiobjective
setting, ESP produces multiple Prescriptors, selected from the Pareto
front of the multiobjective neuroevolution run. The Prescriptor is the
novel aspect of ESP: it makes it possible to discover effective
solutions that do not already exist, even those that might be
overlooked by human decision makers.

In past work, ESP was found to be effective in standard reinforcement
learning benchmarks \cite{francon:gecco20}, as well as in real-world
applications of discovering recipes for growing plants in controlled
environments \cite{johnson:plos19} and in designing e-commerce
websites \cite{miikkulainen:aimag20}.  The benchmarks were useful
because ESP could be evaluated in them as an autonomous decision
making system. However, in real-world applications, including NPI
optimization, it is most naturally used as a system for augmenting
human decision making.  To support this role, it is possible to
include a 'scratchpad' functionality, whereby the human decision maker
can see the predicted outcomes of the ESP-prescribed actions, as well
as modify the prescriptions and weigh the outcomes as part of their
decision making. A scratchpad will be an integral part of the NPI
interactive tool as well (Section~\ref{sc:future}).

Another helpful extension of ESP is to automatically estimate the
confidence in the predicted outcomes. With neural networks, softmax of
the output is often used as an estimation of confidence, but this
measure is often inaccurate \cite{gal:icml16}.  A better approach is to
build a certainty estimation model to complement the Predictor
through a Gaussian process of training data residuals using input and
output kernels  \cite{qiu:iclr20}. This approach will be described in
more detail in Section~\ref{sc:rio}.

In the NPI optimization task, ESP is built to prescribe the NPIs for
the current day such that the number of cases and cost that would
result in the next two weeks is optimized. The details of the Predictor
and Prescriptor in this setup will be described next, after an overview
of the data used to construct them.

\section{Data}

Even though COVID-19 is not the first global pandemic, it is the first
that is recorded in significant detail, providing data that is made
publicly available daily as the pandemic unfolds.

The earliest data that became available included the number of
confirmed cases, the number of deaths, and the number of recovered
patients, per country, region, and day. A well known such source is
the Johns Hopkins University (JHU \cite{jhudata20}), updated daily and
quoted widely in the press. Several predictive models have been built based
on this data, and there is even a Kaggle competition to predict daily
confirmed cases using it \cite{kagglecomp20}. The Kaggle site encourages
companies and organizations to contribute other useful datasets. Other
data is available about e.g.\ the population and medical system in
each country, but they have not yet been used to inform the models.

Once there was enough data to model how the pandemic spreads, an
important question started to emerge: what can we do to contain it?
Pharmaceutical interventions such as treatments and vaccines will take
time to develop, so the focus has been on implementing
non-pharmaceutical interventions, i.e.\ NPIs. The goal is to "flatten
the curve," i.e.\ limit the spread, gain time, and prevent hospitals
from being overwhelmed until a vaccine can be developed
\cite{ferguson:npi1,flaxman:imperial2020}.  Augmenting the health data with
data on NPIs, it would be possible to learn how the NPIs affect health
outcomes.  For instance, since the pandemic affected China and other
parts of Asia before the rest of the world, it would be possible to
learn from their examples.

However, compiling NPI data turned out to be more difficult, and took
longer. Each country took different actions, at different levels, in
different cities or regions. These decision were reported in the
press, but initially were not aggregated and normalized, making it
hard to form a dataset. For instance in the US, such datasets started
to come out only in April
\cite{kaiser:npidata20,keystone:npidata20,covidtracking20}.  Based on
data from the Covid Tracking Project \cite{covidtracking20},
University of Washington's Institute for Health Metrics and Evaluation
(IHME) developed a dashboard that shows the NPI timeline
\cite{ihme20}. Updated continuously, the dashboard shows a projection
for daily deaths and estimated infections (estimated infections are
higher than confirmed infections due to limited testing). The model
includes a social distancing factor computed from the NPIs, held
constant for the future. At a more global scale, Oxford University's
Blavatnik School of Government provides a dataset of the number of
cases, deaths and NPIs for most countries on a daily basis
\cite{hale:data20,oxford:grt20,petherick:working20}. It initially included six
'Closure and Containment' NPIs, but on April 29 was extended to eight
NPIs with more granular intervention levels.  A detailed explanation
of the data is provided in a codebook \cite{oxfordcodebook20}.


\begin{table*}
 \caption{Definitions of the NPI data used in the experiments.
    The data consists of daily 'Closure and containment' measures in the
    'Coronavirus Government Response Tracker' provided by Oxford University
    \cite{hale:data20,petherick:working20,oxfordcodebook20}. There are eight
    kinds of NPIs, each ranging in stringency from 0 (no measures) to
    2, 3, or 4 (full measures). Together with data on daily cases,
    this data is used to train the Predictor model.}
\centering
\resizebox{\textwidth}{!}{%
\begin{tabular}{@{}|l|l|l|l|l|l|@{}}
\toprule
\textbf{NPI name} &
  \textbf{Level 0} &
  \textbf{Level 1} &
  \textbf{Level 2} &
  \textbf{Level 3} &
  \textbf{Level 4} \\ \midrule
C1\_School closing &
  no measures &
  recommend closing &
  \begin{tabular}[c]{@{}l@{}}require closing\\ (only some levels \\ or categories,\\  e.g. just high school,\\  or just public schools)\end{tabular} &
  require closing all levels &
   \\ \midrule
C2\_Workplace closing &
  no measures &
  \begin{tabular}[c]{@{}l@{}}recommend closing\\  (or recommend \\ work from home)\end{tabular} &
  \begin{tabular}[c]{@{}l@{}}require closing\\ (or work from home)\\ for some sectors\\  or categories\\  of workers\end{tabular} &
  \begin{tabular}[c]{@{}l@{}}require closing\\  (or work from home)\\  for all-but-essential\\  workplaces\\  (eg grocery stores,\\  doctors)\end{tabular} &
   \\ \midrule
C3\_Cancel public events &
  no measures &
  recommend cancelling &
  require cancelling &
   &
   \\ \midrule
C4\_Restrictions on gatherings &
  no restrictions &
  \begin{tabular}[c]{@{}l@{}}restrictions on very large\\  gatherings\\ (the limit is above 1000 people)\end{tabular} &
  \begin{tabular}[c]{@{}l@{}}restrictions on \\ gatherings between \\ 101-1000 people\end{tabular} &
  \begin{tabular}[c]{@{}l@{}}restrictions on\\  gatherings between\\  11-100 people\end{tabular} &
  \begin{tabular}[c]{@{}l@{}}restrictions on\\  gatherings of \\ 10 people or less\end{tabular} \\ \midrule
C5\_Close public transport &
  no measures &
  \begin{tabular}[c]{@{}l@{}}recommend closing\\  (or significantly reduce\\  volume/route/means\\  of transport available)\end{tabular} &
  \begin{tabular}[c]{@{}l@{}}require closing\\  (or prohibit most\\  citizens from using it)\end{tabular} &
   &
   \\ \midrule
C6\_Stay at home requirements &
  no measures &
  recommend not leaving house &
  \begin{tabular}[c]{@{}l@{}}require not leaving house\\  with exceptions for daily\\  exercise, grocery shopping,\\  and 'essential' trips\end{tabular} &
  \begin{tabular}[c]{@{}l@{}}require not leaving\\  house with minimal\\  exceptions \\ (e.g.\ allowed to leave\\  once a week, or only\\  one person can leave\\  at a time, etc)\end{tabular} &
   \\ \midrule
\begin{tabular}[c]{@{}l@{}}C7\_Restrictions on\\  internal movement\end{tabular} &
  no measures &
  \begin{tabular}[c]{@{}l@{}}recommend not to\\  travel between regions/cities\end{tabular} &
  \begin{tabular}[c]{@{}l@{}}internal movement \\ restrictions in place\end{tabular} &
   &
   \\ \midrule
C8\_International travel controls &
  no restrictions &
  screening arrivals &
  \begin{tabular}[c]{@{}l@{}}quarantine arrivals\\  from some or all regions\end{tabular} &
  \begin{tabular}[c]{@{}l@{}}ban arrivals from \\ some regions\end{tabular} &
  \begin{tabular}[c]{@{}l@{}}ban on all regions or\\  total border closure\end{tabular} \\ \bottomrule
\end{tabular}
}
\label{tb:npidata}
\end{table*}

The Oxford dataset was used as a source in the current ESP study.  The
models were trained using the 'ConfirmedCases' data for the cases and
'Closure and Containment' data for the NPIs. The other NPI categories in
the dataset, i.e.\ 'Economic response', 'Public Heath' and
'Miscellaneous' measures were not used because they have less direct impact on the
spread of the epidemic. A summary of these NPIs is given in
Table~\ref{tb:npidata}.

The number of cases was selected as the target for the predictions
(instead of number of deaths, which is generally believed to be more
reliable), because case numbers are higher and the data is smoother
overall. The model also utilizes a full 21-day case history which it
can use to uncover structural regularities in the case data. For
instance, it discovers that many fewer cases are reported on the
weekends in France and Spain. However, the data is still noisy for
several reasons:
\begin{itemize}
  \item There are other differences in how cases are reported in each
    country;
  \item Some countries, like the US, do not have a uniform manner of
    reporting the cases;
  \item Cases were simply not detected for a while, and testing policy
    still widely differs from country to country.
  \item Some countries, like China, US, and Italy, implemented NPIs at a
    state / regional level, and it is difficult to express them at the
    country level;
  \item As usual with datasets, there are mistakes, missing days,
    double-counted days, etc.
\end{itemize}
It is also important to note that there is roughly a two-week delay between
the time a person is infected and the time the case is
detected. A similar delay can therefore be expected between the time
an NPI is put in places and its effect on the number of cases.

Despite these challenges, it is possible to use the data to train a
useful model to predict future cases. This data-driven machine
learning approach will be described next.

\section{Data-Driven Predictive Model}
\label{sc:predictive}

With the above data sources, machine learning techniques can be used
to build a predictive model. Good use of recent deep learning
approaches to sequence processing can be made in this process, in
particular recurrent neural networks. However, a method of cascading
the predictions needs to be developed so that they can reach several
steps into the future. Furthermore, methods are needed that keep the predictions within a sensible range even with limited data.

\subsection{Predictor Model Design}
\label{subsec:predictor_design}
This section describes the step-by-step design of the learned predictor.
For a given country, let $x_n$ be the number of new cases on day $n$.
The goal is to predict $x_n$ in the future.
First, consider the minimal epidemic model
\begin{equation}
    x_n = R_nx_{n-1} \implies R_n = \frac{x_n}{x_{n-1}}, \ \ \text{for some} \ \ R_n \geq 0.
\end{equation}
where the factor $R_n$ is to be predicted.
Focusing on such factors is fundamental to epidemiological models, and, when learning a predictive model from data, makes it possible to normalize prediction targets across countries and across time, thereby simplifying the learning task.

Training targets $R_n$ can be constructed directly from daily case data for each country.
However, in many countries case reporting is noisy and unreliable, leading to unreasonably high noise in daily $R_n$.
This effect can be mitigated by instead forming smoothed targets based on a moving average $z_n$ of new cases:
\begin{eqnarray}
    z_n = R_nz_{n-1} \implies R_n = \frac{z_n}{z_{n-1}},\\\nonumber
    \text{\hspace{0pt}where\hspace{5pt}} z_n = \frac{1}{K} \sum_{i=0}^{K-1} x_{n-i}.
\end{eqnarray}
In this paper, $K=7$ for all models, i.e.\ prediction targets are smoothed over the preceding week.

\begin{figure*}
    \centering
    \includegraphics[width=0.75\linewidth]{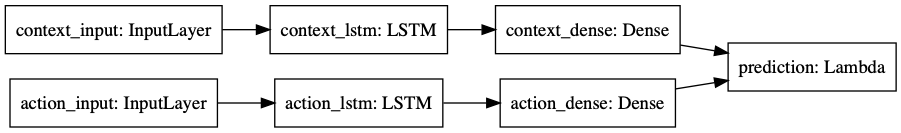}
    \caption{The Predictor Neural Network.
    This diagram shows the Keras representation of the learnable predictor model.
    The previous 21 days of $R_{n-t}$ are fed into the context\_input; the previous 21 days of stringency values for the eight NPIs are fed into the action\_input. 
    The Lambda layer combines the context branch $h$ and the action branch $g$ as specified in Equation~\ref{eq:lambda} to produce a prediction $\hat{R}_n$. The effects of social distancing and endogenous growth rate of the pandemic are processed in separate pathways, making it possible to ensure that stringency has a monotonic effect, resulting in more regular predictions.
    } 
    \label{fg:predictor_diagram}
\end{figure*}

To capture the effects of finite population size and immunity, an additional factor is included that scales predictions by the proportion of the population that could possibly become new cases:
\begin{equation}
    \label{eq:r_definition}
    z_n = \frac{P - y_{n-1}}{P}R_nz_{n-1} \implies R_n = \frac{Pz_n}{(P - y_{n-1})z_{n-1}},
\end{equation}
where $P$ is the population size, and $y_{n} = \sum_{i=0}^{n} x_i$ is the total number of recorded cases by day $n$.
Notice that, when evaluating a trained model, the predicted $\hat{x}_n$ can be recovered from a predicted $\hat{R}_n$ by
\begin{equation}
    \label{eq:convert_to_cases}
    \hat{x}_n = (\hat{R}_n \frac{P - y_{n-1}}{P}- 1) K z_{n-1} + x_{n-K}.
\end{equation}
Note that this formulation assumes that recovered cases are fully immune: When $P = y_{n-1}$, the number of new cases goes to 0. This assumption can be relaxed in the future by adding a factor to Equation~\ref{eq:r_definition} (either taken from the literature or learned) to represent people who were infected and are no longer immune.

The trainable function implementing $\hat{R}_n$ can now be described.
The prediction $\hat{R}_n$ should be a function of (1) NPIs enacted over previous days, and (2) the underlying state of the pandemic distinct from the enacted NPIs.
For the models in this paper, (1) is represented by the NPI restrictiveness values for the past $T=21$ days over all $N=8$ available NPIs, and (2) is represented autoregressively by the $T$ previous values of $R_n$ (or, during forecasting, by the predicted $\hat{R}_n$ when the true $R_n$ is unavailable).
Formally,
\begin{eqnarray}
    &\hat{R}_n = f(\mathbf{A}_n, \mathbf{r}_n), \\\nonumber
    &\text{\hspace{0pt}with\hspace{10pt}} \mathbf{A}_n \in \mathbb{N}_0^{T \times N} \text{\hspace{10pt}and\hspace{10pt}} \mathbf{r}_n \in \mathbb{R}_{\geq0}^{T}.
\end{eqnarray}
In contrast to epidemiological models that make predictions based on today's state only, this data-driven model predicts based on data from the preceding three weeks.

To help the model generalize with a relatively small amount of training data, the model is made more tractable by decomposing $f$ with respect to its inputs:
\begin{eqnarray}
    \label{eq:lambda}
    &\hat{R}_n = f(\mathbf{A}_n, \mathbf{r}_n) = \big(1 - g(\mathbf{A}_n)\big)h(\mathbf{r}_n), \\\nonumber &\text{with} \ \ g(\mathbf{A}_n) \in [0,1] \ \ \text{and} \ \ h(\mathbf{r}_n) \geq 0.
    \end{eqnarray}
Here, the factor $g(\mathbf{A}_n)$ can be viewed as the effect of social distancing (i.e.\ NPIs), and $h(\mathbf{r}_n)$ as the endogenous growth rate of the disease.

To make effective use of the nonlinear and temporal aspects of the data, both $g$ and $h$ are implemented as LSTM models \cite{hochreiter:nc97}, each with a single LSTM layer of 32 units, followed by a dense layer with a single output.
To satisfy their output bounds, the dense layers of $g$ and $h$ are followed by sigmoid and softplus activation, respectively.

Importantly, the factorization of $f$ into $g$ and $h$ makes it possible to explicitly incorporate the constraint that \emph{increasing the stringency of NPIs cannot decrease their effectiveness}.
This idea is incorporated by constraining $g$ to be monotonic with respect to each NPI, i.e.\
\begin{equation}
    \min(\mathbf{A} - \mathbf{A'}) \geq 0 \implies g(\mathbf{A}) \geq g(\mathbf{A'}).
\end{equation}
This constraint is enforced by requiring all trainable parameters of $g$ to be non-negative, except for the single bias parameter in its dense layer.
This non-negativity is implemented by setting all trainable parameters to their absolute value after each parameter update.

Note that although the model is trained only to predict one day in the future, it can make predictions arbitrarily far into the future given a schedule of NPIs by autoregressively feeding the predicted $\hat{R}_{n+t}$ back into the model as input.

For the experiments in this paper, the model for $f$ was implemented in Keras \cite{Chollet:2015}.
The Keras diagram of the model is shown in Figure~\ref{fg:predictor_diagram}.
The model is trained end-to-end to minimize mean absolute error (MAE) with respect to targets $R_n$ using the Adam optimizer \cite{Kingma:14} with default parameters and batch size 32.
MAE was used instead of mean squared error (MSE) because it is more robust to remaining structural noise in the training data.
The last 14 days of data were withheld from the dataset for testing.
For the remaining data, the $R_n$ were clipped to the range $[0, 2]$ to handle extreme outliers, and randomly split into 90\% for training and 10\% for validation during training.
The model was trained until validation MAE did not improve for 20 epochs, at which point the weights yielding the best validation MAE were restored.
Since the model and dataset are relatively small compared to common deep learning datasets, the model is relatively inexpensive to train.
On a 2018 MacBook Pro Laptop with six Intel i7 cores, the model takes $276 \pm 31$ seconds to train (mean and std.\ err.\ computed over 10 independent training runs).

\subsection{Predictor Empirical Evaluation}
\label{subsec:predicor_evaluation}

\begin{table*}[ht]
    \caption{Performance comparison of proposed predictor (NPI-LSTM) with baselines.
    This table shows results along the four metrics described in Section~\ref{subsec:predicor_evaluation} with mean and standard error over 10 trials.
    Interestingly, although RF and SVR do quite well in terms of the loss they were trained on (1-step $\hat{R}_n$ MAE), the simple linear model outperforms them substantially on the metrics that require forecasting beyond a single day, showing the difficulty that off-the-shelf nonlinear methods have in handling such forecasting. In contrast, with the extensions developed specifically for the epidemic modeling case, the NPI-LSTM methods outperforms the baselines on all metrics.
    }
    \centering
    \begin{tabular}{l|r|r|r|r}
    \toprule
        Method & Norm. Case MAE & Raw Case MAE & Mean Rank & 1-step $\hat{R}_n$ MAE \\ \midrule
        MLP & 2.47$\pm 1.22$ & 1089126$\pm 540789$ & 3.19$\pm 0.09$ & 0.769$\pm 0.033$ \\
        RF & 0.95$\pm 0.05$ & 221308$\pm 8717$ & 1.98$\pm 0.10$& 0.512$\pm 0.000$ \\
        SVR & 0.71$\pm 0.00$ & 280731$\pm 0$ & 1.76$\pm 0.09$& 0.520$\pm 0.000$ \\ 
        Linear & 0.64$\pm 0.00$ & 176070$\pm 0$ & 1.63$\pm 0.09$& 0.902$\pm 0.000$ \\
        NPI-LSTM & \textbf{0.42}$\pm 0.04$ & \textbf{154194}$\pm 14593$ & \textbf{1.46}$\pm 0.08$ & \textbf{0.510}$\pm 0.001$ \\
    \bottomrule
    \end{tabular}
    \label{tb:predictor_baseline_comparison}
\end{table*}

To validate the factored monotonic LSTM (NPI-LSTM) predictor design described above, it was compared to a suite of baseline machine learning regression models.
These baselines included linear regression, random forest regression (RF), support vector regression (SVR) with an RBF kernel, and feed-forward neural network regression (MLP).
Each baseline was implemented with sci-kit learn, using their default parameters \cite{scikit-learn}.
Each method was trained independently 10 times on the training dataset described in Section~\ref{subsec:predictor_design}.
The results on the test dataset (last $T^*=14$ days of the $C=20$ countries with the most cases) were evaluated with respect to four complementary performance metrics.
In particular, for the comparisons in this section, training data consisted of data up until May 6, 2020, and test data consisted of data from May 7 through May 20, 2020.

Suppose training data ends on day $n$.
Let $\hat{R}_{n+t}^c$ and $\hat{x}_{n+t}^c$ be the model output and the corresponding predicted new cases (recovered via Equation~\ref{eq:convert_to_cases}) for the $c$th country at day $n+t$.
The metrics were:

\paragraph{1-step $\hat{R}_n$ MAE}
This metric is simply the loss the models were explicitly trained to minimize, i.e.\ minimize $\lvert R_n - \hat{R}_n \rvert$ given the ground truth for the previous 21 days:
\begin{equation}
    \frac{1}{CT^*}\sum_{c=1}^C \sum_{t=1}^{T^*} \left| R_{n+t}^c - \hat{R}_{n+t}^c \right| .
\end{equation}

The remaining three metrics are based not on single-step prediction, but the complete 14 day forecast for each country:

\paragraph{Raw Case MAE}
This is the most intuitive metric, included as an interpretable reference point.
It is simply the MAE w.r.t.\ new cases over the 14 test days summed over all 20 test countries:
\begin{equation}
    \sum_{c=1}^C \left| \sum_{t=1}^{T^*} x_{n+t}^c - \sum_{t=1}^{T^*} \hat{x}_{n+t}^c  \right| .
\end{equation}

\paragraph{Normalized Case MAE}
This metric normalizes the case MAE of each country by the number of true cases in the 14 day window, so that errors are in a similar range across countries.
Such normalization is important for aggregating results over countries that have different population sizes, or are in different stages of the pandemic:
\begin{equation}
    \frac{1}{C} \sum_{c=1}^C \frac{\left| \sum_{t=1}^{T^*} x_{n+t}^c - \sum_{t=1}^{T^*} \hat{x}_{n+t}^c  \right|}{\sum_{t=1}^{T^*} x_{n+t}^c} .
\end{equation}

\paragraph{Mean Rank}
This metric ranks the methods in terms of case error for each country, and then averages over countries.
It indicates how often a method will be preferred over others on a country-by-country basis:
\begin{equation}
    \frac{1}{C} \sum_{c=1}^C \text{rank}\Bigg(\left| \sum_{t=1}^{T^*} x_{n+t}^c - \sum_{t=1}^{T^*} \hat{x}_{n+t}^c  \right|\Bigg),
\end{equation}
where $\text{rank}(\cdot)$ returns the rank of the error across all five methods, i.e.\ the method with the lowest error receives rank of 0, the next-best method receives rank of 1, and so on.
\\

Of these four metrics, Normalized Case MAE gives the most complete picture of how well a method is doing, since it combines detailed case information of Raw Case MAE with fairness across countries similar to Mean Rank. 
The results are shown in Table~\ref{tb:predictor_baseline_comparison}.
NPI-LSTM outperforms the baselines on all metrics.
Interestingly, although RF and SVR do quite well in terms of the loss on which they were trained (1-step $\hat{R}_n$ MAE), the simple linear model outperforms them substantially on the metrics that require forecasting beyond a single day, showing the difficulty that off-the-shelf nonlinear methods have in handling such forecasting.

\begin{figure*}
    \centering
    \includegraphics[width=0.45\textwidth]{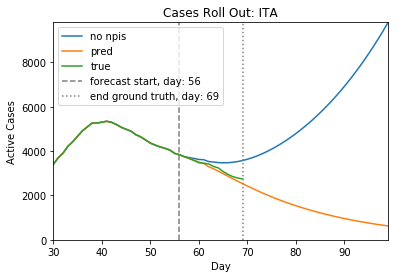}
    \hfill
    \includegraphics[width=0.45\textwidth]{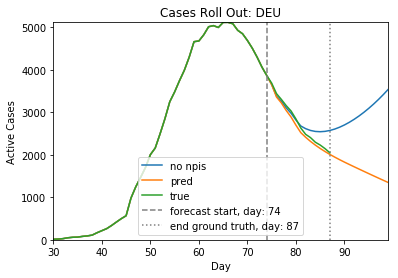}\\
    \includegraphics[width=0.45\textwidth]{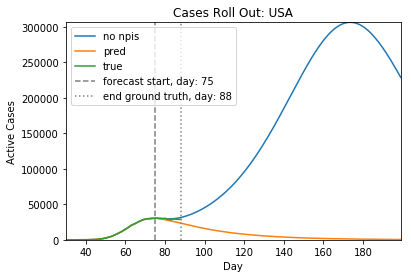}
    \hfill
    \includegraphics[width=0.45\textwidth]{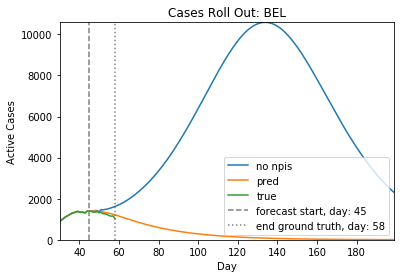}
    \caption{Illustrating the predictive ability of the NPI-LSTM          
    model. Actual and projected cases are shown for four sample countries. The model predicts the number of cases accurately for the first 14 days where it can be compared with the actual future
    data (between the vertical lines). The prolonged 180-day predictions are also meaningful,
    reducing the number of cases to zero with stringent NPIs, and
    predicting a major increase followed by an eventual decrease with
    less stringent NPIs. Thus, with proper constraints, data-driven
    machine learning models can be surprisingly accurate in predicting
    the progress of the pandemic even with limited data.}
    \label{fg:predictorexamples}
\end{figure*}

To verify that the predictions are meaningful and accurate, four
example scenarios, i.e.\ four different countries at different stages
of the pandemic, are plotted in Figure~\ref{fg:predictorexamples} (active cases at each day is approximated as the sum of new cases over the prior 14 days).
Day 0 represents the point in time when 10 total cases were diagnosed; in
each case, stringent NPIs were enacted soon after. The predictor was trained on data up until April 17, 2020, and the predictions
started on April 18, with 21 days of data before the start day given to the predictor as initial input. Assuming the NPIs in
effect on the start day will remain unchanged, it will then predict
the number of cases 180 days into the future. Importantly, during the first 14
days its predictions can be compared to the actual number of
cases. For comparison, another prediction plot is generated from the
same start date assuming no NPIs from that date on. As
can be seen from the figure, (1) the predictions match the actual
progress well, (2) assuming the current stringent NPIs continue, the
cases will eventually go to 0, and (3) with no NPIs, there is a
large increase of cases, followed by an	eventual decrease as the
population becomes immune. The predictions thus follow meaningful
trajectories.

The main conclusion from these experiments is that the data-driven
approach works surprisingly well, even with limited data. As will be
discussed in Section~\ref{sc:future}, it should	be possible to improve
it in the future, when more and	better data becomes available. It
might also be possible to combine the strengths	of the different
approaches to prediction and epidemiological modeling. In any case,
the current predictive model already makes	it possible to build
Prescriptors for ESP, as will be discussed in Section~\ref{sc:prescriptor}.
The way confidence can be estimated in it will be described next.

\subsection{Modeling Uncertainty in Predictions}
\label{sc:rio}

An important aspect of any decision system is to estimate confidence
in its outcomes. In prescribing NPIs, this means estimating uncertainty
in the Predictor, i.e.\ deriving confidence intervals on the predicted
number of future cases. In simulation models such as those reviewed in
Section~\ref{sc:background}, variation is usually created by running the
models multiple times with slightly different initial conditions or
parameter values, and measuring the resulting variance in the
predictions. With neural network predictors, it is possible to measure
uncertainty more directly by combining a Bayesian model with it
\cite{neal:book96,garnelo:icml18,kim:arxi19}. Such extended models tend
to be less accurate than pure predictive models, and also harder to set
up and train \cite{gal:icml16,lakshminarayanan:nips17}.

A recent alternative is to train a separate model to estimate
uncertainty in point-prediction models \cite{qiu:iclr20}. In this
approach, called RIO, a Gaussian Process is fit to the original residual errors in the training set. The I/O kernel of RIO utilizes both input and output of the original model so that information can be used where it is most reliable. In several benchmarks, RIO has been shown to construct reliable confidence intervals. Surprisingly, it can then be used to improve the point predictions of the original model, by correcting them towards the estimated mean. RIO can be directly applied to any machine learning model without modifications or retraining. It therefore forms a good basis for estimating uncertainty in the COVID-19 Predictor as well.

In order to extend RIO to time-series predictions, the hidden states of the two LSTM models (before the lambda layer in Figure~\ref{fg:predictor_diagram}) are concatenated and fed into the input kernel of RIO. The original predictions of the predictor are used by the output kernel. RIO is then trained to fit the residuals of the original predictions. During deployment, the trained RIO model then provides a Gaussian distribution for the calibrated predictions. The details of this process are presented in Algorithm~\ref{app:pseudo_RIO}.

\begin{algorithm*}[tb]
	\caption{Procedure of applying RIO to the COVID-19 Predictor}
	\label{app:pseudo_RIO}
	\begin{algorithmic}[1]
		\REQUIRE ${}$
		\\ $\{((\mathbf{A}_i, \mathbf{r}_i),R_i)\}_{i=1}^n$: training data
		\\ $\mathbf{\hat{R}}=\{\hat{R}_i\}_{i=1}^n$: original predictions on training data
		\\ $(\mathbf{A}_*, \mathbf{r}_*)$: data to be predicted
		\\ $\hat{R}_*$: original prediction on $(\mathbf{A}_*, \mathbf{r}_*)$
		\ENSURE ${}$
		\\ $\hat{R}^{\prime}_*\sim \mathcal{N}(\hat{R}_*+\bar{\hat{e}}_*, \mathrm{var}(\hat{e}_*))$: a distribution of calibrated prediction\\
		{\bf \hspace{-23pt} Training Phase:}
		\STATE Calculate residuals $\mathbf{e}=\{e_i = R_i - \hat{R}_i\}_{i=1}^n$
		\STATE Obtain hidden states of the two LSTM models in original predictor $\{g(\mathbf{A}_n), h(\mathbf{r}_n)\}_{i=1}^n$
		\FOR{each optimizer step}
		\STATE Calculate covariance matrix $\mathbf{K}_c((\{g(\mathbf{A}_n), h(\mathbf{r}_n)\}_{i=1}^n,\mathbf{\hat{R}}), (\{g(\mathbf{A}_n), h(\mathbf{r}_n)\}_{i=1}^n, \mathbf{\hat{R}}))$,\\ where each entry is given by $k_c(((g(\mathbf{A}_i), h(\mathbf{r}_i)),\hat{R}_i), ((g(\mathbf{A}_j), h(\mathbf{r}_j)),\hat{R}_j))$\\ = $k_\mathrm{in}((g(\mathbf{A}_i), h(\mathbf{r}_i)), (g(\mathbf{A}_j), h(\mathbf{r}_j))) + k_\mathrm{out}(\hat{R}_i, \hat{R}_j),\hspace{5pt} \mathrm{for}\hspace{3pt} i,j=1,2,\ldots,n$
		\STATE Optimize GP hyperparameters by maximizing log marginal likelihood\\ $\log p(\mathbf{e}|\{g(\mathbf{A}_n), h(\mathbf{r}_n)\}_{i=1}^n,\mathbf{\hat{R}}) = -\frac{1}{2}\mathbf{e}^\top(\mathbf{K}_c(\cdot)+\sigma_n^2\mathbf{I})^{-1}\mathbf{e}-\frac{1}{2}\log|\mathbf{K}_c(\cdot)+\sigma_n^2\mathbf{I}|-\frac{n}{2}\log 2\pi$
		\ENDFOR
		{\bf \hspace{-23pt} Deployment Phase:}
		\STATE Calculate residual mean $\bar{\hat{e}}_* =\mathbf{k}_*^\top(\mathbf{K}_c(\cdot)+\sigma_n^2\mathbf{I})^{-1}\mathbf{e}$ and residual variance\\ $\mathrm{var}(\hat{e}_*) =k_c(((g(\mathbf{A}_*), h(\mathbf{r}_*)),\hat{R}_*), ((g(\mathbf{A}_*), h(\mathbf{r}_*)),\hat{R}_*))-\mathbf{k}_*^\top(\mathbf{K}_c(\cdot)+\sigma_n^2\mathbf{I})^{-1}\mathbf{k}_*$
		\STATE Return distribution of calibrated prediction $\hat{R}^{\prime}_*\sim \mathcal{N}(\hat{R}_*+\bar{\hat{e}}_*, \mathrm{var}(\hat{e}_*))$
	\end{algorithmic}
\end{algorithm*}

\begin{table*}[ht]
	\scriptsize
	\caption{\label{tab:rio} Results after applying RIO to predictor}
	\centering
	\begin{tabular}{l c c c c c}
		\toprule
		Dataset & original MAE & MAE with RIO & 95\% CI Coverage & 90\% CI Coverage & 68\% CI Coverage \\ \hline
		Training dataset & 0.0319 & 0.0312 & 0.952 & 0.921 & 0.756 \\
		Testing dataset & 0.0338 & 0.0337 & 0.929 & 0.899 & 0.710 \\
		\bottomrule
	\end{tabular}

CI coverage means the percentage of testing outcomes that are within the estimated confidence intervals (CIs).
\end{table*}

To validate this process empirically with COVID-19 data, the data was preprocessed in four steps: (1) Among the 30 most affected countries in terms of cases, those with the most accurate predictions were selected, resulting in 17 countries with MAE less than 0.04. (2) The outlier days that had an $R$ larger than 2.0 were removed from the data. (3) The earliest 10 days (after the first 21 days) were removed as well, focusing training on more recent data. (4) For each country, 14 days were selected randomly as the testing data, and all the remaining days were used as the training data. The hyperparameters in these steps were found to be appropriate empirically.  Table~\ref{tab:rio} shows the results. The conclusion is that RIO constructs reasonable confidence intervals at several confidence levels, and slightly improves the prediction accuracy. It can therefore be expected to work well in estimating confidence in the NPI prescription outcomes as well.

However, RIO will first need to be extended to model uncertainty in time series. Because NPI-LSTM forecasts are highly nonlinear and autoregressive, analytic methods are intractable. Instead, given that the predictor model with RIO returns both the mean and the quartiles for $\hat{R}_n$, the quartiles after $t$ days in the future can be estimated via Monte Carlo rollouts. Specifically, for each step in each rollout, instead of predicting $\hat{R}$ and feeding it back into the model to predict the next step, $\hat{R}$ is sampled from the Gaussian distribution returned by RIO, and this sample is fed back into the model. Thus, after $T^*$ steps, a sample is generated from the forecast distribution. Given several such samples (100 in the experiments in this paper), the upper and lower quartile are computed empirically for all forecasted days $1 \leq t \leq T^*$.

Thus, RIO makes it possible to estimate uncertainty in the
predictions, which in turn helps the decision maker interpret and
trust the results, i.e.\ how reliable the outcomes are for the
recommendations that the Prescriptors generate.  The method for
discovering good Prescriptors will be described next.

\section{Evolutionary Prescriptive Model}
\label{sc:prescriptor}

Whereas many different models could be used as a Predictor, the
Prescriptor is the heart of the ESP approach, and needs to be
constructed using modern search techniques. This section describes the
process of evolving neural networks for this task. A number of example
strategies are presented from the Pareto front, representing trade-offs between
objectives, as well as examples for countries at different stages of the
pandemic, and counterfactual examples comparing possible vs.\ actual
outcomes. General conclusions are drawn
on which NPIs matter the most, and how they could be implemented most
effectively.

\begin{figure*}
    \centering
    \includegraphics[width=0.75\textwidth]{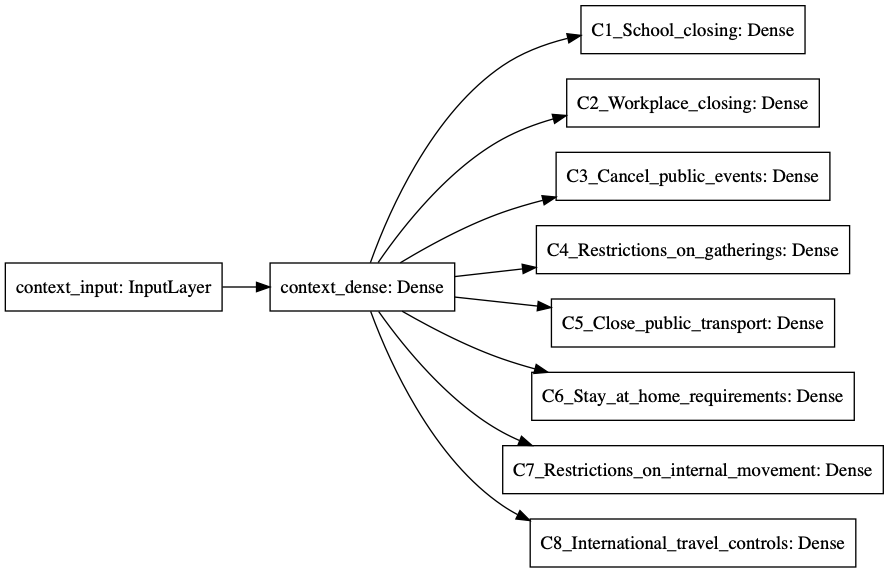}
    \caption{The Prescriptor Neural Network. Given 21 past days 
      of case information ($R_{n-t}$ in Equation~\ref{eq:r_definition})                    
      as input (context\_input), the network
      generates recommended stringency values for each of the eight
      NPIs. The network is fully connected with one hidden layer.
      Because there are no targets, i.e.\ the optimal NPIs are not
      known, gradient descent cannot be used; instead, all weights and
      biases are evolved based on how well the network's NPI
      recommendations work along the cases and cost objectives, as
      predicted by the Predictor.}
    \label{fg:prescriptormodel}
\end{figure*}

Any of the existing neuroevolution methods \cite{stanley:naturemi19}
could be used to construct the Prescriptor as long as it evolves the
entire network including all of its weight parameters: Neural
architecture search cannot be used easily since there are no targets
(i.e.\ known optimal NPIs) with which to train it with gradient
descent.  The most straightforward approach of evolving a vector of
weights for a fixed topology is therefore used and found to be sufficient in
this case.  The Prescriptor model (Figure~\ref{fg:prescriptormodel})
is a neural network with one input layer of size 21, corresponding to 
case information $R_{n-t}$ (as defined in Equation~\ref{eq:r_definition}) 
for the prior 21 days. This
input is the same as the context\_input of the Predictor. The input
layer is followed by a fully-connected hidden layer of size 32 with 
the tanh activation function,
and eight outputs (of size one) with the sigmoid activation function. The
outputs represent the eight possible NPIs, which will then be input to
the Predictor.  Each output is further scaled and rounded to represent the
corresponding NPI stringency levels: within [0,2] for 'Cancel public events',
'Close public transport', and 'Restrictions on internal movement';
[0,3] for 'School closing', 'Workplace closing', and 'Stay at home';
[0,4] for 'Restrictions on gatherings' and 'International travel controls'.

The initial population uses orthogonal initialization of weights in
each layer with a mean of 0 and a standard deviation of 1
\cite{saxe2013orthogonalinit}. The population size is 250 and the top
6\% of the population is carried over as elites. Parents are selected
by tournament selection of the top 20\% of candidates using the
NSGA-II algorithm \cite{deb:ieeetec02}. Recombination is
performed by uniform crossover at the weight-level, and there is a
20\% probability of multiplying each weight by a mutation factor,
where mutation factors are drawn from $\mathcal{N}(1, 0.1)$.

Prescriptor candidates are evaluated according to two objectives: (1)
the expected number of cases according to the prescribed NPIs, and 
(2) the total stringency of the prescribed NPIs (i.e.\ the sum of 
the stringency levels of the eight NPIs), serving as a proxy
for their economic cost. Both measures are averaged over the next 
180 days and over the 20 countries with the
most deaths in the historical data, which at the time of the experiment
were United States, United Kingdom,
Italy, France, Spain, Brazil, Belgium, Germany, Iran, Canada,
Netherlands, Mexico, China, Turkey, Sweden, India, Ecuador, Russia,
Peru, Switzerland. Both objectives have to be minimized.

On the evaluation start date, each Prescriptor is fed with the last 
21 days of case information. Its outputs
are used as the NPIs at the evaluation start date, and combined with
the NPIs for the previous 20 days. These 21 days of case information
and NPIs are given to the Predictor as input, and it outputs the
predicted case information for the next day.  This
output is used as the most recent input for the next day, and the
process continues for the next 180 days. At the end of the process,
the average number of predicted new cases over the 180-day period 
is used as the value of the first objective. Similarly, the
average of daily stringencies of the prescribed NPIs over the 180-day
period is used as the value for the second objective. 

\begin{figure}
    \centering
    \includegraphics[width=0.6\textwidth]{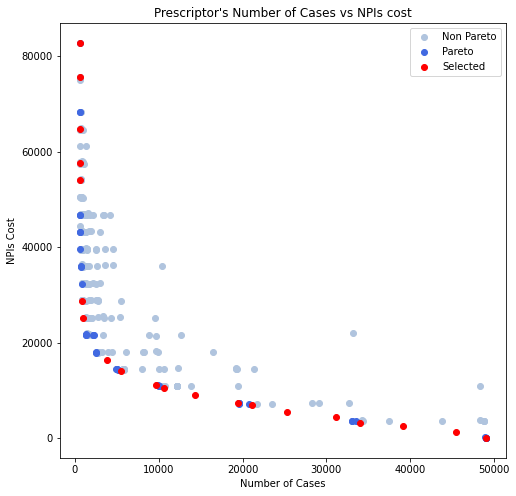}
    \caption{Fitness of the Final Population along the Case and           
        Cost Objectives. The candidates at the lower left side are on          
      the Pareto front, representing the best tradeoffs. Those
      in red are used in the examples below and in the interactive            
      demo (numbered 0 to 19 from left to right). They are the 20 
      candidates with the highest crowding distance in NSGA-II. 
      The other candidates in the Pareto front are
      in dark blue and other final population candidates
      in light blue. An animation of
      how this population evolved can be seen at          
      \url{https://evolution.ml/esp/npi}.}
    \label{fg:pareto}
\end{figure}

After each candidate is evaluated in this manner, the next generation
of candidates is generated. Evolution is run for 110 generations, or
approximately 72 hours, on a single CPU host. During the course of
evolution, candidates are discovered that are increasingly more fit
along the two objectives.  In the end, the collection of candidates
that represent best possible tradeoffs between objectives (the Pareto
front, i.e.\ the set of candidates that are better than all other
candidates in at least one objective) is the final result of the
experiment (Figure~\ref{fg:pareto}).  From this collection, it is up
to the human decision maker to pick the tradeoff that achieves a
desirable balance between cases and cost. Or put in another way, given a
desired balance, the ESP system will find the best to achieve it
(i.e.\ with the lowest cost and the lowest number of cases).

\begin{figure*}
    \centering
    \begin{minipage}{0.49\textwidth}
    \centering
    \includegraphics[width=\textwidth]{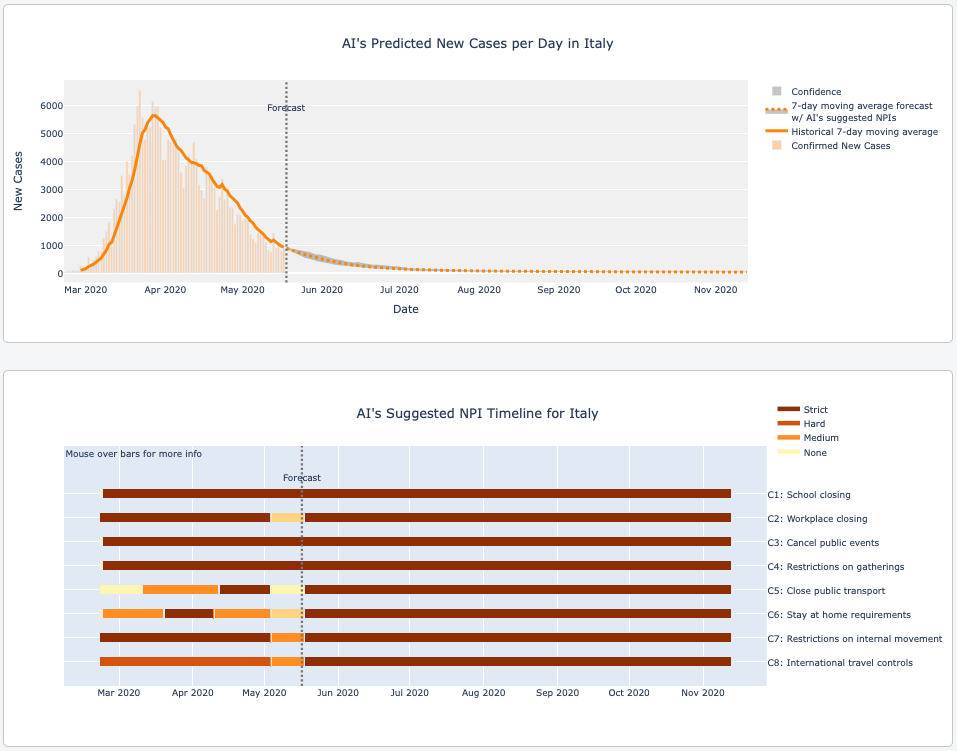}\\
    {\footnotesize $(a)$ Prescriptor~0:\\[-1ex]Minimize cases}
    \end{minipage}
    \hfill
    \begin{minipage}{0.49\textwidth}
    \centering
    \includegraphics[width=\textwidth]{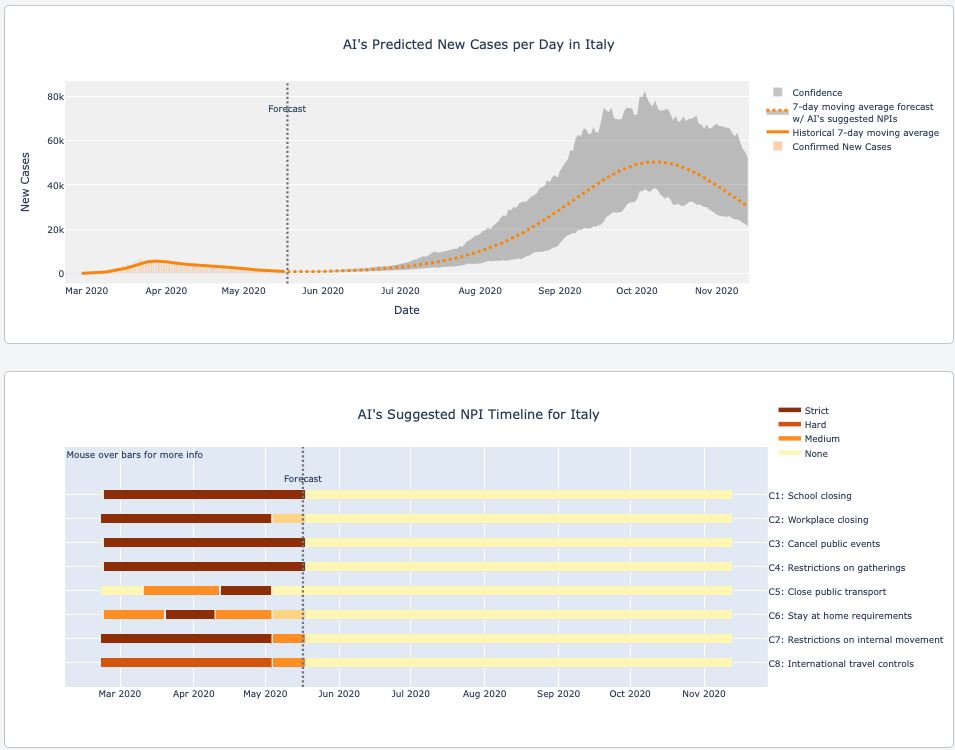}\\
    {\footnotesize $(b)$ Prescriptor~19:\\[-1ex]Minimize Stringency}
    \end{minipage}\\[2ex]
    \begin{minipage}{0.49\textwidth}
    \centering
    \includegraphics[width=\textwidth]{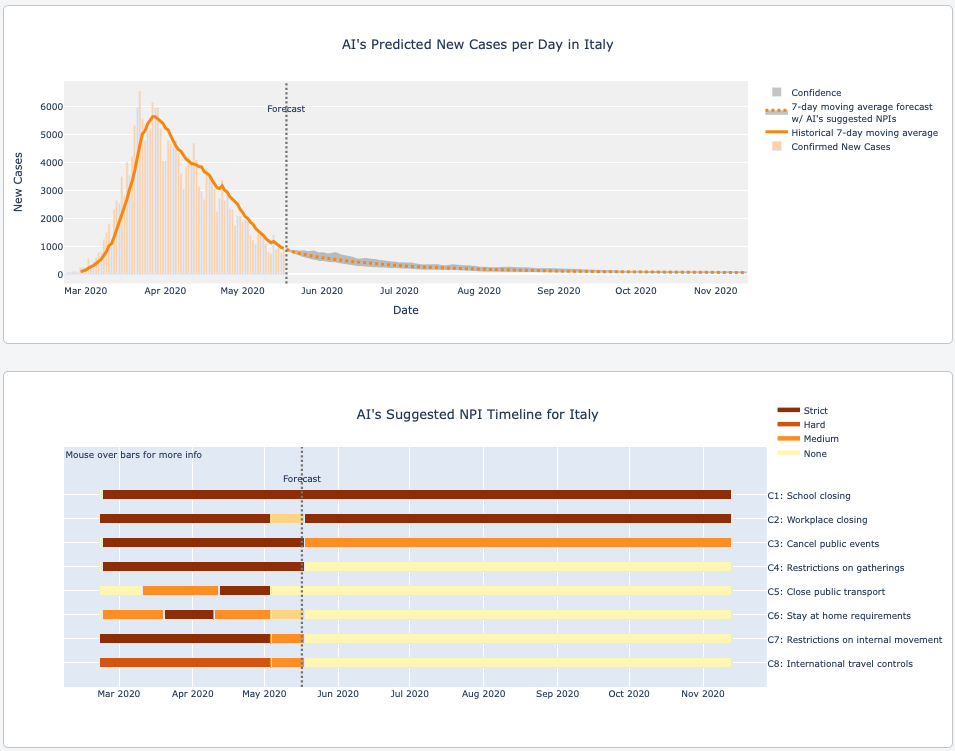}\\
    {\footnotesize $(c)$ Prescriptor~6:\\[-1ex]An example of stringent tradeoff}
    \end{minipage}
    \begin{minipage}{0.49\textwidth}
    \centering
    \includegraphics[width=\textwidth]{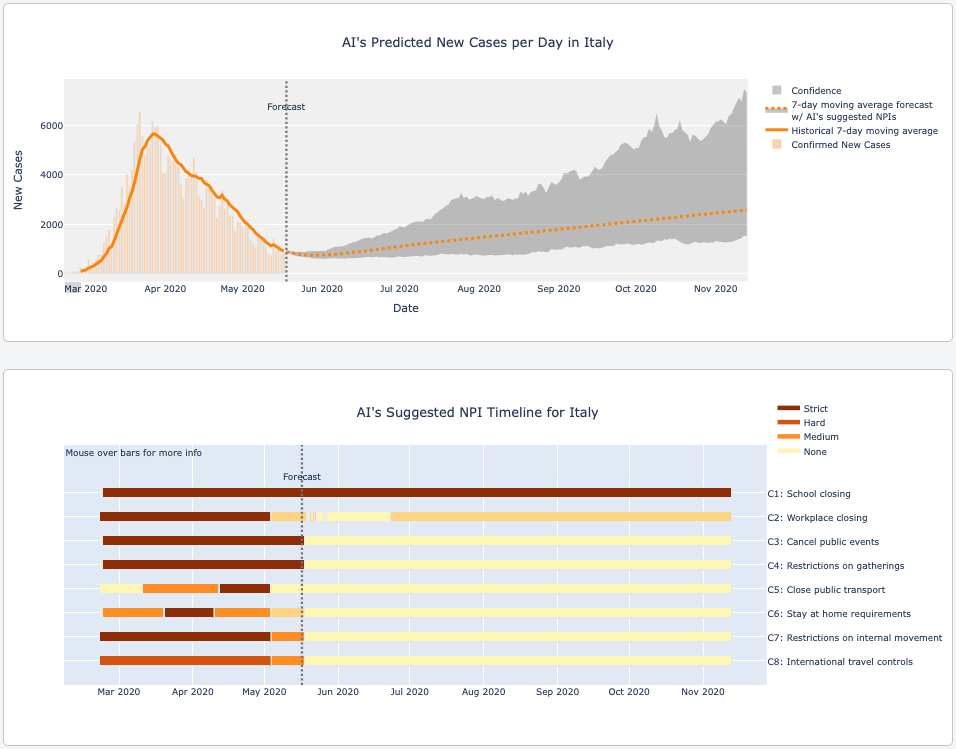}\\
    {\footnotesize $(d)$ Prescriptor~8:\\[-1ex]An example of non-stringent tradeoff}
    \end{minipage}\\
    \caption{Comparison of Different Prescriptors from the Pareto Front.
    The recommendations of four different Prescriptors are shown for Italy.
    Daily cases are shown as orange vertical bars and their seven-day moving average
    as the orange line. The vertical line indicates the start of the forecast,
    and the gray area represents uncertainty around the prediction. The NPI prescriptions are
    shown below the case plot as horizontal bars, with color representing stringency.
    $(a)$ The Prescriptor that minimizes the number of cases                    
    recommends a full set of NPIs at their maximum level of stringency.         
    $(b)$ The Prescriptor that minimizes the NPI stringency recommends          
    lifting all NPIs, which is likely to result in a high number of cases.
    $(c)$ A Prescriptor that tries to minimize the number of cases              
    while lifting as many NPIs as possible recommends keeping                   
    restrictions mostly on schools and workplaces.
    $(d$) A Prescriptor that tries to reduce the cost more by opening
    up workplaces completely may result in cases climbing up. 
    In this manner, the human decision maker can              
    explore the tradeoffs between cases and cost, and the ESP system            
    will recommend the best ways to achieve it.                                 
    }
    \label{fig:results_italy}
\end{figure*}

To illustrate these different tradeoffs,
Figure~\ref{fig:results_italy} shows the NPI Presprictions and the
resulting forecasts for four different Prescriptors from the Pareto
front for one country, Italy, on May 18th, 2020. The Prescriptor that minimizes cases
prescribes the most stringent NPIs across the board, and as a result,
the number of cases is minimized effectively. The Prescriptor that
minimizes NPI stringency lifts all NPIs right away, and the number of
cases is predicted to explode as a result. The third Prescriptor was chosen from the
middle of the Pareto front, and it represents one particular way to
balance the two objectives. It lifts most of the NPIs, allows some
public events, and keeps the schools and workplaces closed. As a
result, the number of cases is still minimized, albeit slightly slower
than in the most stringent case.  Lifting
more of the NPIs, in particular workplace restrictions, is likely to cause
the number of cases to start climbing. In this manner, the decision maker
may explore the Pareto front, finding a point that achieves the most
desirable balance of cases and cost for the current stage of the pandemic.

\begin{figure*}
    \centering
    \begin{minipage}{0.49\textwidth}
    \centering
    \includegraphics[width=\textwidth]{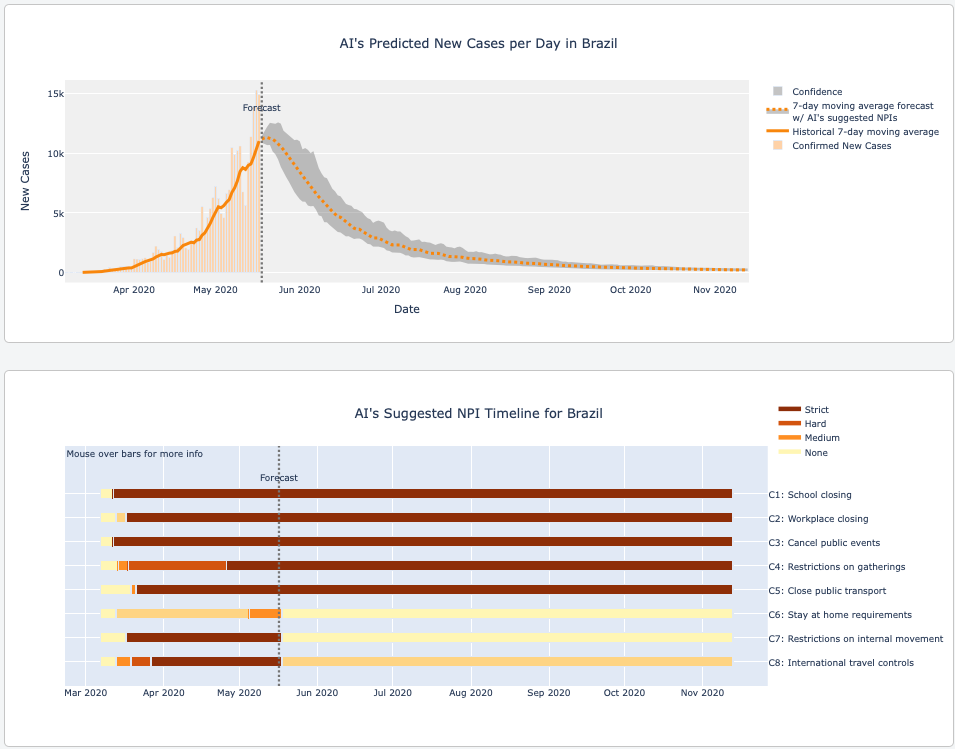}\\
    {\footnotesize $(a)$ Rapid Expansion:\\[-1ex]Prescriptor~4}
    \end{minipage}
    \hfill
    \begin{minipage}{0.49\textwidth}
    \centering
    \includegraphics[width=\textwidth]{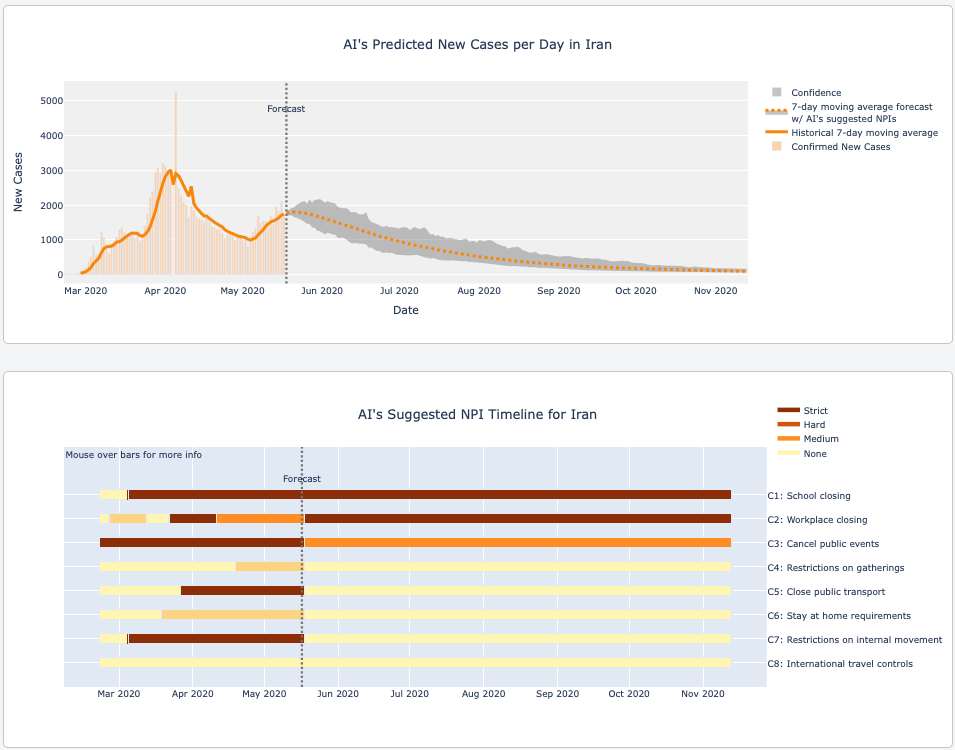}\\
    {\footnotesize $(b)$ Second Wave\\[-1ex]Prescriptor~6}
    \end{minipage}\\[2ex]
    \begin{minipage}{0.49\textwidth}
    \centering
    \includegraphics[width=\textwidth]{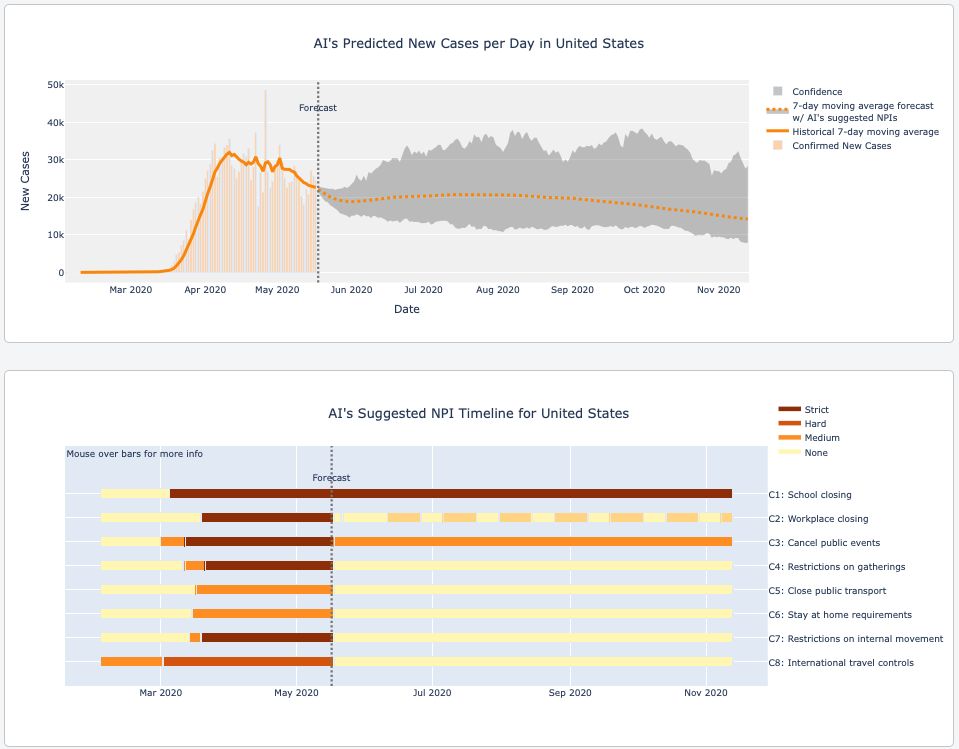}\\
    {\footnotesize $(c)$ Relatively Flat\\[-1ex]Prescriptor~7}
    \end{minipage}
    \begin{minipage}{0.49\textwidth}
    \centering
    \includegraphics[width=\textwidth]{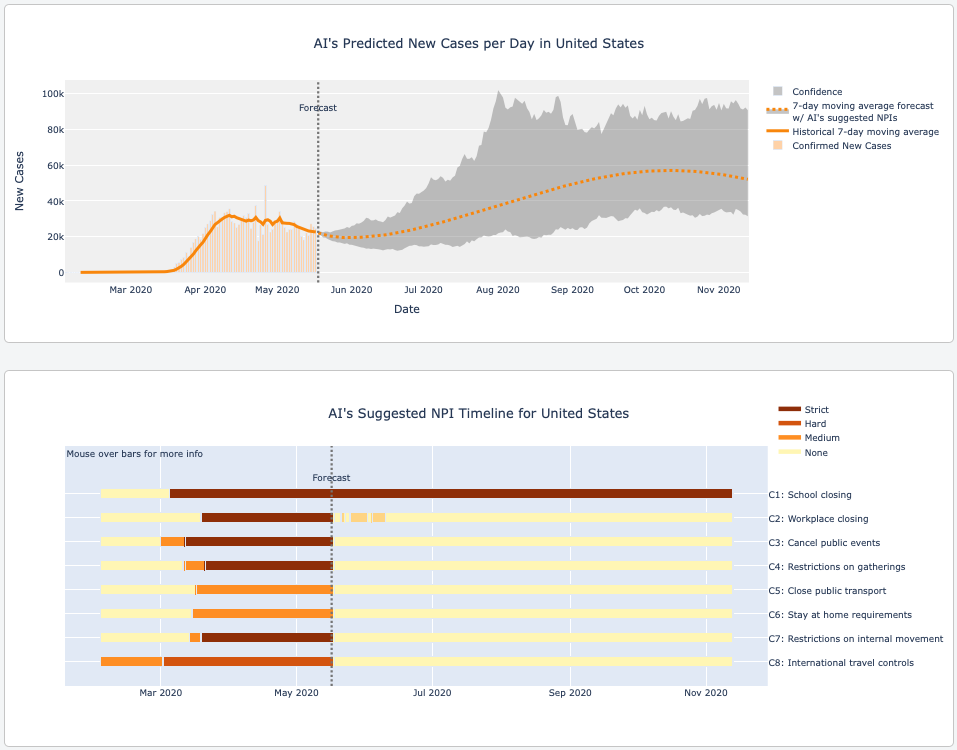}\\
    {\footnotesize $(d)$ Relatively Flat\\[-1ex]Prescriptor~9}
    \end{minipage}
    \caption{Comparison of Tradeoff Prescriptions for Countries           
        at Different Stages of the Pandemic.  The Prescriptors chosen          
      represent a midrange in the balance between cases and cost,               
      similar to that of Figure~\ref{fig:results_italy}$(c)$.  $(a)$ For          
      Brazil, where the pandemic is spreading rapidly at this point, Prescriptor~4
      minimizes cases effectively while allowing some freedom of movement.
      $(b)$ For Iran, where the pandemic appears to be entering a second wave,
      a more stringent Prescriptor~6 strikes a similar balance.
      $(c)$ For the US, where cases            
      are relatively flat at this point, a less stringent Prescriptor~7 allows
      reducing cases gradually with minimal cost.  $(d)$ In contrast, an even slightly less
      stringent Prescriptor such as~9 would allow a high number of cases to return.
      Interestingly, in all these cases as             
      well as in Figure~\ref{fig:results_italy}$(c)$, schools and                 
      workplaces are subject to restrictions while others are lifted. Also,            
      Prescriptor~7 often includes an alternation of stringency             
      levels, suggesting a way to reduce the cost of the NPI while potentially
      keeping it effective. Thus, evolution discovers where NPIs may            
      have the largest impact, and can suggest creative ways of                 
      implementing them.  }
    \label{fg:stages}
\end{figure*}

The shadowed area in Figures~\ref{fig:results_italy}$(a)$-$(d)$ represents the uncertainty of the prediction, i.e., areas between 25th and 75th percentiles of the 100 Monte Carlo rollouts under uncertainty estimated through RIO. The width of the shadowed area is normalized to match the scale of the forecasts (dotted line). It is often asymmetric because there is more variance in how the pandemic can spread than how it can be contained. Whereas uncertainty is narrow with stringent Prescriptors (Figure~\ref{fig:results_italy}$(a)$,$(c)$), it often increases significantly with time with less stringent ones. The increase can be especially dramatic with Prescriptors with minimal NPIs, such as those in Figures~\ref{fig:results_italy}$(b)$ and~$(d)$. The reason is that at the time these forecasts were made, not much training data existed yet about this stage of the pandemic (i.e.\ the stage where countries are lifting most NPIs after the peak of the pandemic has passed). The model's suggestions at this point should be taken as indicative only; with more training data in the future, these confidence estimates are likely to become more reliable. However, this result can already be interpreted to suggest that such minimal-NPI prescriptions are fragile, making the country vulnerable to subsequent waves of the pandemic (see also Figures~\ref{fg:stages}$(c)$ and~$(d)$).

To illustrate this process, Figure~\ref{fg:stages} shows
possible choices for three different countries at
different stages of the pandemic on May 18th, 2020. For
Brazil, where the pandemic is still spreading rapidly at this point,
a relatively stringent Prescriptor~4 allows some freedom of movement
without increasing the cases much compared to full lockdown. For
US, where the number of cases has been relatively flat,
a less stringent Prescriptor~7 may be chosen, limiting restrictions 
to schools, workplaces, and public events. However, if NPIs are lifted
too much, e.g.\ by opening up the workplaces and allowing public events, high numbers of cases
are predicted to return.  For Iran, where there is
a danger of a second wave, Prescriptor~6 provides more stringent NPIs to prevent cases from
increasing, still limiting the restrictions to schools, workplaces and public events.

Interestingly, across several countries	at different stages of the
pandemic, a consistent pattern emerges: in order to keep the number of
cases flat, other NPIs can be lifted gradually, but workplace and
school restrictions need to be in effect much longer. Indeed these are
the two	activities where people	spend a	lot of time with other people
indoors, where it is possible to be exposed to significant amounts of
the virus \cite{kay:quillette20,park:emerindis20,lu:emerindis20}. In other activities, such as gatherings and
travel, they may come to contact with many people briefly and often
outdoors, mitigating the risk. Therefore, the main conclusion that can
already be drawn from these prescription experiments is
that it	is not the casual contacts but the extended contacts that
matter. Consequently, when planning for	lifting	NPIs, attention	should
be paid	in particular to how workplaces and schools can be	opened
safely.

Another interesting conclusion can be drawn from Figure~\ref{fg:stages}(c):
Alternating between weeks of opening workplaces and partially closing them
may be an effective way to lessen the impact on the economy while reducing cases.
This solution is interesting because it shows how evolution	can be
creative and find surprising and unusual solutions that are
nevertheless effective.	There is of course much	literature documenting
similar surprising discoveries in computational evolution
\cite{johnson:plos19,miikkulainen:aimag20,lehman:alife20}, but it is
encouraging to see that they are possible also in the NPI
optimization domain.  While on/off alternation of school and workplace
closings may sound unwieldy, it is a real possibility \cite{chowdhury:eurjourepi20}.
Note also that it is the only creative solution
available to the Prescriptor: there are no options in its output for
e.g.\ alternating
remote and in-person work, extending school to wider hours,
improving ventilation, moving classes outside, requiring masks, or
other ways of possibly reducing exposure. How to best implement such distancing at school	and workplace
is left	for human decision makers at this point; the model, however,
makes a	suggestion that	coming up with such solutions may make it
possible to lift the NPIs gradually, and thereby avoid secondary waves
of cases.

Thus, in the early stages, the ESP approach may suggest how to ``flatten the           
curve'', i.e.\ what NPIs should be implemented in order to slow down
the spread of the disease. At later stages, it may recommend
how the NPIs can be lifted and the economy restarted safely. 
A third role for the approach is to go back in time and evaluate counterfactuals,
i.e.\ how well NPI
strategies other than those actually implemented could have worked. In this
manner, it may be possible to draw conclusions not only about the accuracy
and limitations of the modeling approach, but also lessons for future
waves of the current pandemic, for new regions where it is still spreading,
as well as for future pandemics.

\begin{figure*}
    \centering
    \begin{minipage}{0.49\textwidth}
    \centering
    \includegraphics[width=\textwidth]{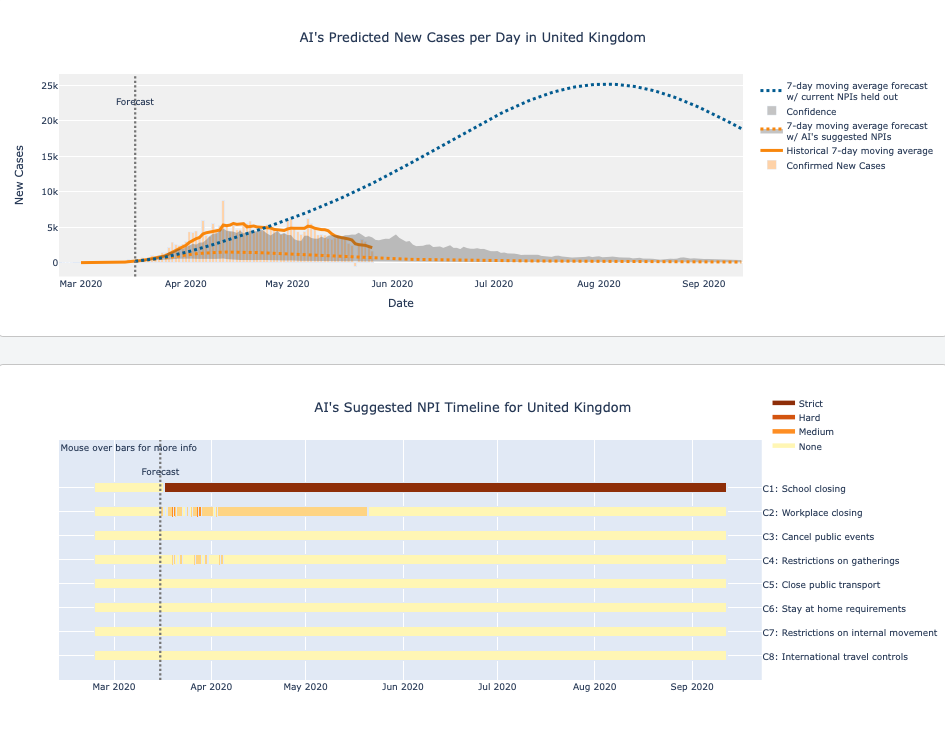}
    {\footnotesize $(a)$ Earlier NPIs in the UK\\[-1ex]Prescriptor~8}
    \end{minipage}
    \hfill
    \begin{minipage}{0.49\textwidth}
    \centering
    \includegraphics[width=\textwidth]{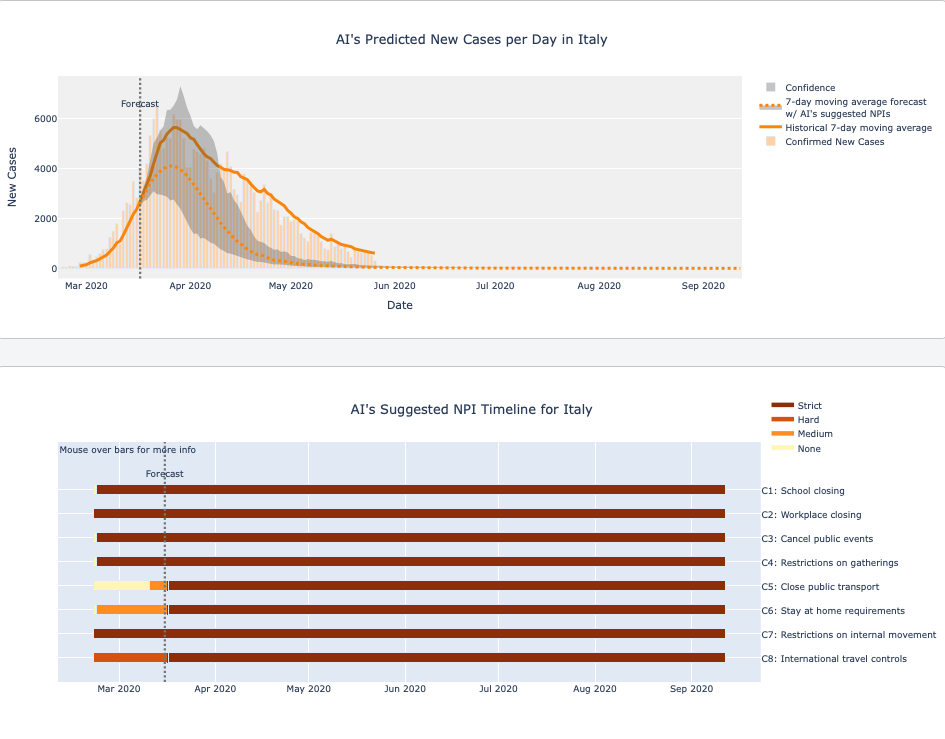}
    {\footnotesize $(b)$ Predicted vs.\ actual in Italy\\[-1ex]Prescriptor~0}
    \end{minipage}
    \caption{Evaluating the Model with Counterfactuals.
      Going back in time to make prescriptions makes it possible
      to evaluate how accurate the model is and to draw lessons
      for the remainder of this pandemic and future pandemics.
      $(a)$ After an initial phase of mild NPIs, a lockdown in the 
      UK averted a sharp rise in cases. However,
      Prescriptor~8 would have recommended earlier NPIs that 
      could have resulted in an even better outcome without a full lockdown.
      (Uncertainty estimates are unreliable in this case and not shown.)
      $(b)$ In Italy, a full lockdown in effect on March 16th should have resulted in 
      much fewer cases than it did, suggesting that cultural and
      demographic factors were different than in other countries,
      and that the implementation of NPIs need to take such
      factors into account in order to be effective.}
    \label{fg:counterfactuals}
\end{figure*}

For instance in the UK on March 16th, the NPIs actually in effect were 
the mild 'recommend work from home' and 'recommend cancel public events'.
With only these NPIs, the predicted number of cases could have been quite
high (Figure~\ref{fg:counterfactuals}$(a)$). A lockdown was
implemented on March 24th, and the actual case numbers were significantly smaller. However,
it is remarkable that Prescriptor~8 would have required closing schools 
already on March 16th, and the predicted cases could have been much fewer
even without a more extensive lockdown. Thus, the model suggests that an early
intervention is crucial, and indeed other models have been used to draw
similar conclusions \cite{pei:medrxiv20}.
What is interesting is that ESP suggests that it may be possible to
control the pandemic with less than full lockdowns if acted early enough.
Of course, the fully trained model was not available at that point, however
these lessons may still be useful for countries and regions that
are still in early stages, as well as for future pandemics.

Some of the limitations of the data-driven approach also become evident in retrospective 
studies. For instance Italy, where the pandemic took hold before most of the rest of the world,
was supposed to be in a lockdown on March 16th (which started already on February 23rd). Yet, 
the model predicts that under such a lockdown (suggested e.g.\ by Prescriptor~0 for that date),
the number of cases should have been
considerably smaller than they actually were (Figure~\ref{fg:counterfactuals}$(b)$). 
The uncertainty is wide but the model's prediction is remarkably different from those of 
many other countries. Part of the reason may
be that the population in Italy did not adhere stringently to the NPIs
at that point; after all, the full scale of the pandemic was not
yet evident. The population may also have been older and more
susceptible than elsewhere. The data used to train the model comes from 20 different
countries and at a later stage of the pandemic spread, and these populations
may have followed social distancing more carefully---therefore, the model's predictions
on the effectiveness of lockdowns may be too optimistic for Italy. Even with 
the uncertainty, this result suggests that local factors like culture, economy,
population demographics, density, and mobility, may need to be taken into account 
in the models. It also suggests that the implementation of NPIs
need to be sensitive to such factors in order to be effective.

Retrospective studies also show that the epidemic needs to be well
underway for the predictions to be reliable. For instance,
Italy only had 0 to three cases per day until February 22, when they
jumped to 17 in one single day.  Trying to predict the pandemic e.g.\ on
March 1st does not lead accurate results. Much of the spread of the virus 
at that stage may be due to unpredictable superspreader events,
like a church gathering, choir practice, conference,
or a soccer match \cite{frieden:emerindis20}.

Overall, however, the data-driven ESP approach works surprisingly well
even with the current limited data, and can be a useful tool in
understanding and dealing with the pandemic. An interactive demo, 
available on the web, that makes it possible to
explore prescriptions and outcomes of the ESP model like those reviewed
in this section, will be described next.

\section{Interactive Demo}
\label{sc:demo}

\begin{figure*}
    \centering
    \includegraphics[width=0.8\textwidth]{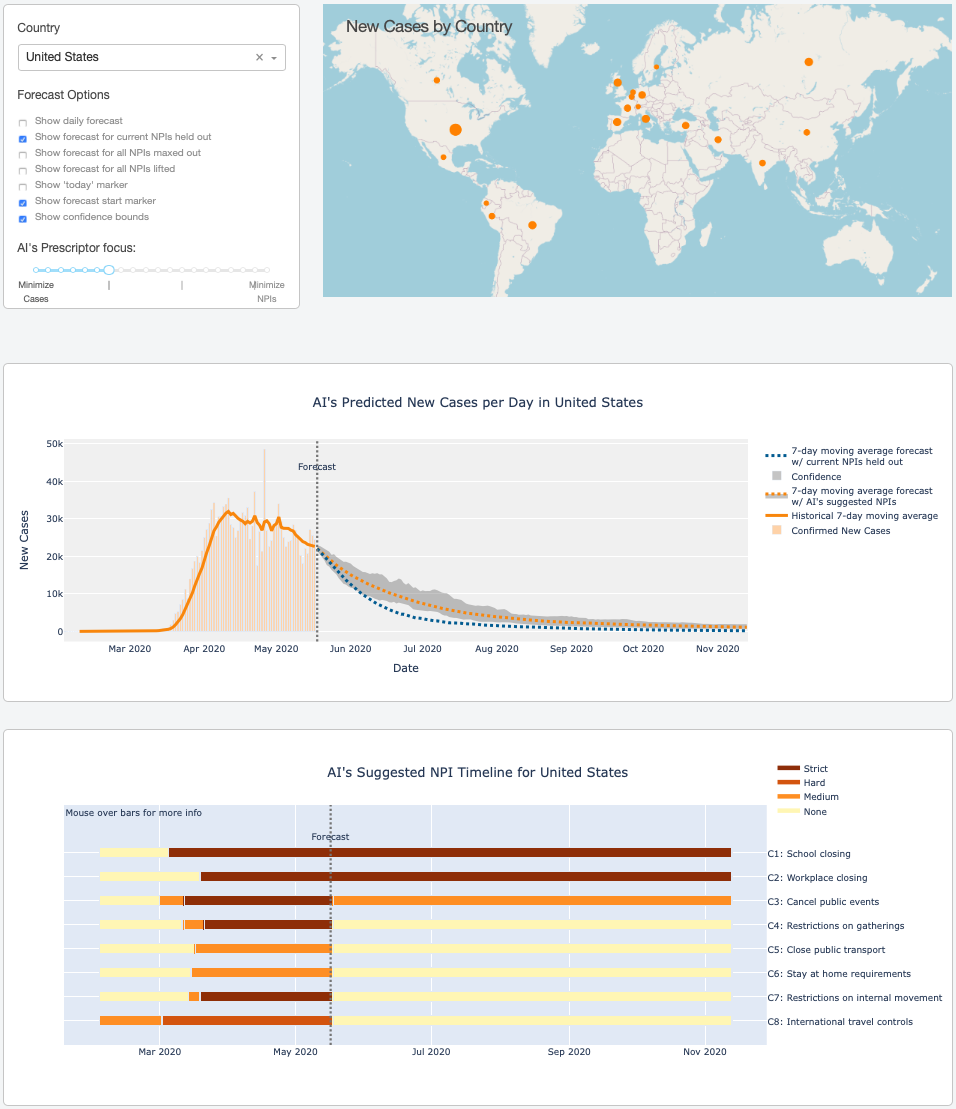}
    \caption{An Interactive Demo of ESP in the NPI Optimization           
        Problem. Selecting the country from the map and a Prescriptor          
      with the desired case/cost trade-off from the Pareto front                 
      slider, the demo shows the predicted cases as a plot over time,           
      and the prescribed NPIs with color-coded stringency over                  
      time. The demo can currently be used to understand the potential          
      for the approach, and possibly in the future to augment human             
      decision making in the pandemic. The demo is available at
      \url{https://evolution.ml/esp/npi}.}
    \label{fg:demo}
\end{figure*}

To help understand the mechanisms and possibilities of ESP models, an
interactive demo of the current state of the approach to NPI
optimization is available at \url{https://evolution.ml/esp/npi}
(Figure~\ref{fg:demo}). This demo will change as the models improve and
new functionality is added.\footnote{The examples presented in this paper   
can be replicated with the appropriate snapshot of the demo:                
\url{https://evolution.ml/demos/npidashboard/?forecast\_folder=20200523\_000001}
for Figures~\ref{fig:results_italy},~\ref{fg:stages}, and~\ref{fg:demo}, and \url{https://evolution.ml/demos/npidashboard/?forecast\_folder=20200317\_000002}
for Figure~\ref{fg:counterfactuals}.}. At the time of this writing, the
following interactions are  possible:

The user can select a country by clicking on the map, and a
Prescriptor from the Pareto front by clicking on the slider between
Cases and NPIs. At the very left, the Presciptors prefer to minimize
cases and therefore usually recommend establishing nearly all possible
NPIs. At the very right, the Prescriptors prefer to minimize NPIs and
therefore usually recommend lifting nearly all of them---usually
resulting in an explosion of cases. The most interesting Prescriptors
are therefore somewhere in the middle-left of this range. Some of them
are able to keep the cases flat while lifting most of the NPIs, as
was discussed in Section~\ref{sc:prescriptor}.

The cases are plotted over time in the middle of the page. The bars
indicate past history, used to initialize the Predictor, and future
predictions as a line plot with confidence bounds towards the right. 
The prescribed NPIs are
shown over time in the chart at the bottom of the page, with darker
colors indicating more stringent version of each NPI. For the
Prescriptors that balance the cases and number of NPIs, it is often
possible to see an alternating pattern of stringency over time.

With the demo it is possible to	explore	the options for	different
countries at different stages of the pandemic. However,	it is
important to keep in mind that the demo	is only	a demonstration	of the
potential of the ESP approach: With current limited data it is not yet
possible to make reliable recommendations in a particular case and
especially with less stringent NPIs. In the
aggregate, it is possible to draw general conclusions, as was done
above. With more and better data and further development, the demo may
eventually develop into a tool that can be	used to	augment human decision	making
in the pandemic.

\section{Future Work}
\label{sc:future}

Given the encouraging results in this paper, the most compelling
direction of future work consists of updating the model with new data
as it becomes available. As NPIs are gradually lifted in many
countries, the volume of data will increase, but data will also be
more relevant for making decisions in the future. COVID-19 testing
will hopefully improve as well so that the outcome measures will be
more reliable. The models can be extended to predicting and minimizing
deaths as well as cases. Such a multi-task learning environment should
make predictions in each task more accurate \cite{liang:gecco18}. 
Instead of the current eight NPIs with a few stringency
levels, data on more fine-grained and detailed NPIs may become
available, as well as data on more fine-grained locations,
such as US counties. In other words, data will improve in volume,
relevance, accuracy, and detail, all of which will help make the
predictors more precise, and thereby improve prescriptions.

Technically the most compelling direction is to take advantage of
multiple prediction models, in particular more traditional
compartmental or network models reviewed in
Section~\ref{sc:background}. General assumptions about the spread of
the disease are built in to these models, and they can thus serve as a
stable reference when data is otherwise lacking in a particular case.
On the other hand, it is sometimes hard to estimate the parameters
that these models require, and data-driven models can be more mode
accurate in specific cases. A particularly good approach might be to
form an ensemble from these models (as is often done in machine
learning; \cite{zaremba:arxiv14,miikkulainen:agebook18}), and
thereby combine their predictions systematically to maximize accuracy.

Another way to make the system more accurate and useful is to improve
the outcome measures. Currently the cost of the NPIs is
proxied based on how many of them are implemented and at what
stringency level. Economic impact is difficult to measure, and the
current approach, however approximate, already works relatively well.
However, it may be possible to develop more accurate measures
based on a variety of economic indicators, such as unemployment,
consumer spending, and GNP. They need to be developed for each country
separately, given different social and economic structures. With 
such measures, ESP would be free to find
surprising solutions that, while stringent, may not have as high an
economic impact.

The retrospective example of Italy in Figure~\ref{fg:counterfactuals}$(b)$
suggests that it may be difficult to transfer conclusions from one country
to another, and to make accurate recommendations early on in the epidemic. An
important future analysis will be to analyze systematically how much data 
and what kind of data is necessary. For instance, if the model had been developed based on the
data from China, would it have worked for Iran and
Italy? Or after China, Iran, and Italy, would it have worked for the US
and the rest of Europe? That is, how much transfer is there between countries
and how many scenarios does the model need to see before it becomes reliable?
The lessons can be useful for the rest of the COVID-19 pandemic, as well as
for future pandemics.

An important aspect of any decision system is to make it trustworthy,
i.e.\ estimate confidence in its decisions and predictions, allow
users to utilize their expert knowledge and explore alternatives, and
explain the decision recommendations. The first step was already taken
in this study by applying the RIO uncertainty
estimation method (Section~\ref{sc:rio}) to the predictions. This approach
may be improved in the future
by grouping the countries according to original predictor performance, then training a dedicated RIO model for each group. In this way, each RIO model focuses on learning the predictive uncertainty of countries with similar patterns, so that the estimated confidence intervals become more reliable. As a further step, the estimated uncertainty can be used by the Prescriptor to make safer decisions.

Second, a prescription ``scratchpad'' can be included, allowing the
user to not only see the prescription details (as shown in
Section~\ref{sc:demo}), but also modify them by hand. In this manner,
before any prescriptions are deployed, the user can utilize expert
knowledge that may not be available for ESP. For instance, some NPIs in
some countries may not be feasible or enforceable at a given time. The
interface makes it possible to explore alternatives, and see the
resulting outcome predictions immediately. In this manner, the user
may find more refined prescriptions than those proposed by ESP, or
convince him/herself that they are unlikely to exist.

Third, currently the prescriptions are generated by an evolved neural
network, which may perform well in the task, but does not provide an
explanation of how and why it arrived at a given prescription. In the
future, it may be possible to evolve explicit rule sets for this task
as well, or instead \cite{shahrzad:gptp19,hodjat:gptp18}. Rule sets
are readable, specifying which feature values in the context lead to
which prescriptions. They can therefore be used to generate
explanations of the learned decision strategies, and thereby make it
easier for human decision makers to understand and trust them.

While the current ESP model of NPI optimization is a promising
demonstration, if enough better data becomes available in the next few
months, it may be possible to use the tool during the current COVID-19
pandemic to make useful recommendations. The general approach can also
be developed further in the long term, eventually allowing decision
makers to minimize the impact of future pandemics.

\section{Conclusion}

Recent advances in AI have made it possible to not only predict what
would happen, but also prescribe what should be done. Also, widely
available data has recently made it possible to build data-driven
models that are surprisingly accurate in their predictions. This paper
puts these two advances together to derive recommendations for NPIs in
the current COVID-19 crisis. While preliminary, the model already
leads to insights in which NPIs are most important to get right, as
well as how they might be implemented most effectively.  With further
data and development, the approach may become a useful tool for policy
makers,	helping	them to	minimize impact	of the current as well as
future pandemics.

\bibliographystyle{myplain}
\bibliography{refs}

\end{document}